\documentclass{article}
\usepackage[most]{tcolorbox}

\usepackage{fontspec}

\newfontfamily\thaifont[
  Scale=MatchLowercase,
  Path=./fonts/,
  Extension=.ttf,
  UprightFont=NotoSerifThai-Regular
]{NotoSerifThai-Regular}

\newfontfamily\amharicfont[
  Scale=MatchLowercase,
  Path=./fonts/,
  Extension=.ttf,
  UprightFont=NotoSerifEthiopic-Regular
]{NotoSerifEthiopic-Regular}

 \usepackage[preprint]{neurips_2026}

\usepackage[utf8]{inputenc} 
\usepackage[T1]{fontenc}    
\usepackage{hyperref}       
\usepackage{url}            
\usepackage{booktabs}       
\usepackage{amsfonts}       
\usepackage{nicefrac}       
\usepackage{microtype}      
\usepackage{xcolor}         

\usepackage{enumitem}
\usepackage{graphicx}
\usepackage{microtype}
\usepackage{subcaption}
\usepackage{booktabs} 
\usepackage{listings}
\usepackage{amsmath}
\usepackage{booktabs}
\usepackage{multirow}
\usepackage[table]{xcolor}
\usepackage{makecell}
\usepackage{tcolorbox}
\usepackage{placeins}
\usepackage{float}
\usepackage{wrapfig}
\definecolor{careblue}{RGB}{232,242,255}

\newcommand{\careppo}[1]{\textsc{CARE-PPO}#1}
\newcommand{\nutri}[1]{\textsc{NutriBench}#1}

\newtcolorbox[auto counter, number within=section]{promptbox}[2][]{
  title=Box~\thetcbcounter: #2,
  colback=blue!3,
  colframe=blue!40,
  fonttitle=\bfseries,
  breakable,
  #1
}

\newtcolorbox[auto counter, number within=section]{examplebox}[2][]{
  title=Example~\thetcbcounter: #2,
  colback=cyan!1,
  colframe=cyan!50!black,
  fonttitle=\bfseries,
  breakable,
  #1
}

\title{From Critic to Confidence: PPO for Language-Based Quantitative Prediction with Confidence Estimation}

\author{
Mehak Dhaliwal \quad
Rasta Tadayon \quad
Andong Hua \quad
Haewon Jeong \quad
Yao Qin \\[0.75em]
University of California, Santa Barbara \\[0.25em]
\texttt{mdhaliwal@ucsb.edu, yaoqin@ucsb.edu}
}

\begin{document}

\maketitle

\begin{abstract}
LLMs can perform language-based quantitative prediction from unstructured inputs, but remain susceptible to hallucinations and overconfident errors, making it critical to know not only \emph{what} a model predicts, but \emph{when} its predictions can be trusted. We introduce \careppo{}, a reinforcement learning framework that establishes a connection between loss prediction for uncertainty estimation and actor--critic PPO fine-tuning, enabling joint learning of accurate numerical estimates and reliable confidence signals in language-based quantitative prediction. \careppo{} uses a \textbf{C}onfidence-\textbf{A}ligned \textbf{R}eward for \textbf{E}stimation, defined as a function of prediction error, to provide dense error-aware feedback to the actor while inducing the critic to learn a value function aligned with prediction quality. During inference, we repurpose the critic as a confidence estimator. Across two real-world tasks in healthcare and finance and two Qwen-3 model scales (4B and 8B), \careppo{} achieves strong quantitative prediction performance, while producing significantly better-aligned confidence estimates through the critic than logit-based and verbalized baselines. These gains persist under realistic out-of-distribution settings across domains, spanning linguistic and domain shifts. Finally, CARE-PPO reduces task-specific overfitting on general instruction-following prompts, consistent with the broader generalization advantages of RL fine-tuning over supervised approaches.

\end{abstract}

\section{Introduction}
Recent advances in large language models (LLMs) have shown that they are not only powerful language generators, but also excel in quantitative prediction tasks across domains such as healthcare~\citep{hua2024nutribench}, finance~\citep{vedula2025quantile}, and scientific analysis~\citep{lewkowycz2022solving}. However, LLMs remain susceptible to hallucinations and overconfident errors~\citep{achiam2023gpt}, a limitation that is especially concerning in high-stakes domains such as healthcare and law~\citep{dahl2024large, omar2025multi}. For instance, LLMs show promise for accurate carbohydrate estimation from meal descriptions~\citep{hua2024nutribench}, a task critical for deciding insulin dosage in patients with diabetes, where overconfident but incorrect carbohydrate estimates can lead to severe glycemic consequences. Deploying such models in practice therefore requires more than accurate point estimates — users in safety-critical settings need to know not only~\emph{what} the model predicts, but also~\emph{when to trust} that prediction in order for humans to intervene when necessary~\citep{kalai2025language, tao2024trust}.

While a growing body of work studies confidence estimation for LLMs in classification and open-ended generation tasks~\citep{geng2024survey}, it remains largely underexplored for numerical prediction in language-based settings, where existing work predominantly focuses on producing accurate point estimates without considering reliable confidence signals~\citep{tang2024understanding, jacobs2024regression, zausinger2024regress, vacareanu2024words}. To address this gap, we study language-based quantitative prediction with confidence estimation, where an LLM generates numerical estimates through natural language reasoning, accompanied by confidence signals that reflect prediction reliability.

\begin{figure*}[t]
    \centering
    \includegraphics[width=0.95\textwidth]{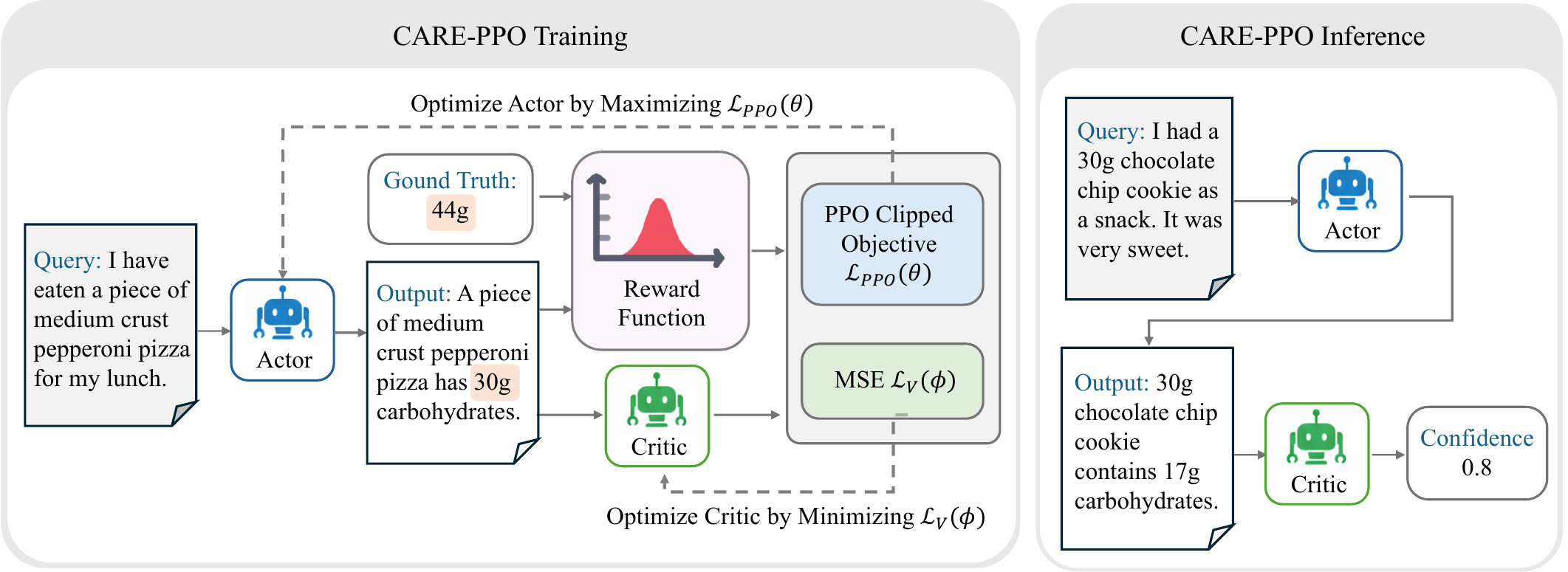}
    \caption{Overview of \careppo{} (\textbf{C}onfidence-\textbf{A}ligned \textbf{R}eward for \textbf{E}stimation with PPO). \textbf{Left:} During training, the actor generates a numerical prediction from the input, which is evaluated using a confidence-aligned reward defined as a function of prediction error. The policy is optimized with PPO using GAE, while the critic is trained to regress to the empirical return. \textbf{Right:} At inference time, the critic’s value estimate is repurposed as a confidence score for the actor’s prediction.}
    \label{fig:pipeline}
\end{figure*}

Specifically, we draw a novel connection between loss prediction for uncertainty estimation~\citep{gollakota2025does, lahlou2021deup} and actor--critic reinforcement learning fine-tuning for LLMs~\citep{schulman2017proximal, ouyang2022training}. Prior work characterizes a model's predictive uncertainty at a given input as the expected loss incurred by the predictor at that point, making estimation of this quantity a loss prediction problem~\citep{gollakota2025does, lahlou2021deup}. In other words, identifying when a prediction is unreliable corresponds to identifying when the model is likely to incur high error.
This perspective naturally connects to the actor--critic framework of PPO~\citep{schulman2017proximal}, where the critic learns a value function that estimates the expected return from a given state (Equation~\ref{eqn:value_loss}): when the reward is defined as a function of prediction error, the critic's value function converges to an estimate of the actor's expected error — realizing loss prediction as a natural byproduct of RL training rather than a post-hoc procedure.

Building on this, we introduce \careppo{} (\textbf{C}onfidence-\textbf{A}ligned \textbf{R}eward for \textbf{E}stimation with PPO), a reinforcement learning framework for language-based quantitative prediction that jointly trains an actor to generate accurate numerical estimates through natural language reasoning and a critic whose value function serves as a confidence estimator aligned with prediction error. We design the reward as a monotonic function of prediction error, serving a dual purpose. First, it provides a dense, error-aware training signal that captures degrees of correctness and guides the actor toward accurate numerical predictions. Second, because the critic is trained to predict returns under this reward, it learns a value function aligned with prediction quality: high values correspond to low expected error, while low values correspond to high expected error. This allows the critic to naturally serve as a confidence estimator without requiring explicit confidence supervision. 

We illustrate this framework in Figure~\ref{fig:pipeline}, and summarize our main contributions below.

\begin{itemize}[leftmargin=*]
    \item \textbf{Problem and insight.} We study language-based quantitative prediction with confidence estimation and establish a connection between loss prediction theory and actor--critic RL fine-tuning: when the reward is aligned with prediction error, the PPO critic naturally learns to estimate the actor's expected loss, enabling reliable confidence estimation without explicit supervision.

    \item \textbf{\careppo{}.} We propose \careppo{}, a PPO-based actor--critic framework with a Confidence-Aligned Reward for Estimation. By defining the reward as a monotonic function of prediction error, \careppo{} provides dense error-aware feedback to the actor while inducing the critic to serve as an error-aligned confidence estimator at inference time.

     \item \textbf{Comprehensive evaluation.} We evaluate \careppo{} on two real-world quantitative prediction tasks spanning healthcare and finance across Qwen-3 4B and 8B models~\citep{yang2025qwen3}, studying both Gaussian and linear reward formulations. \careppo{} achieves competitive prediction performance while substantially improving confidence alignment over logit-based and verbalized baselines.

    \item \textbf{Robustness and task awareness.} We show that \careppo{}'s confidence-alignment gains persist under realistic out-of-distribution settings, including linguistic and domain shifts. We further show that \careppo{} reduces task-specific overfitting on general instruction-following queries, helping preserve general-purpose behavior beyond the target prediction setting.
\end{itemize}

\section{Related Work}

\paragraph{LLMs for language-based quantitative prediction.}
Prior work on applying LLMs to language-based quantitative prediction has primarily followed three directions: (i) using LLM-derived representations as fixed input features for downstream models~\citep{imperial2021bert, tang2024understanding}, (ii) adapting LLMs through task-specific fine-tuning to directly produce numerical predictions~\citep{jacobs2024regression, zausinger2024regress}, and (iii) employing in-context learning to perform zero-shot or few-shot numerical estimation from natural-language prompts~\citep{vacareanu2024words, gopali2025context}. While feature-based approaches can be effective, they treat the LLM primarily as an encoder and thus do not leverage its generative interface for interactive use or natural-language explanations. Meanwhile, both fine-tuning and in-context learning emphasize point estimates rather than rich language generation. In this study, we instead focus on fine-tuning LLMs for language-based quantitative prediction in a way that preserves their generative and conversational capabilities for transparency and interaction.

\paragraph{Confidence estimation in LLMs.}
The growing adoption of LLMs has increased the need for reliable confidence estimates aligned with answer correctness. 
Prior work can be broadly categorized into \textit{black-box} and \textit{white-box} approaches, depending on whether internal model states are accessed~\citep{geng2024survey}. 
 
\textit{Black-box} approaches infer confidence using only model outputs. These include confidence verbalization methods~\citep{lin2022teaching, xiong2023can, tian2023just, yang2024verbalized}, which often suffer from systematic overconfidence~\citep{wei2022chain, mei2025reasoning, kirichenko2025abstentionbench}, as well as sampling-based techniques that estimate confidence from output variability using consensus heuristics~\citep{kang2025scalable}, semantic diversity~\citep{aichberger2025improving}, or entropy-based measures~\citep{kuhn2023semantic}. \textit{White-box} methods instead leverage internal signals such as token probabilities, entropy, or likelihood shifts~\citep{gupta2024language, duan2024shifting, huang2023look, kadavath2022language}, or train probes over hidden representations to predict correctness probability~\citep{azaria2023internal, mielke2022reducing}. Despite their differences, most approaches infer confidence post hoc rather than optimizing it during training. 
In contrast,~\careppo{} explicitly trains a critic to align predicted confidence with quantitative prediction error, enabling confidence estimates that are directly optimized during training rather than inferred after the fact.

\paragraph{Confidence-aware LLM fine-tuning.}
At the intersection of confidence estimation and fine-tuning for language models, prior work primarily follows two directions. One line of work incorporates confidence or uncertainty signals as auxiliary rewards to improve downstream task performance, including speaker–listener frameworks~\citep{stengel2024lacie} and reinforcement learning methods that derive rewards from self-confidence~\citep{li2025confidence, he2025beyond}. A second line of work directly fine-tunes models to output calibrated confidence or uncertainty, with approaches including SFT methods that optimize verbalized confidence using tokenized Brier score losses~\citep{li2025conftuner}, as well as methods that leverage human preference signals to guide confidence calibration~\citep{tao2024trust}. These methods inherit the limitations of SFT, which can suffer from ``catastrophic forgetting'' and degradation of prior capabilities~\citep{luo2025empirical}.
In contrast, our~\careppo{} framework jointly optimizes for correctness and confidence calibration, explicitly rewarding accurate predictions while learning well-calibrated uncertainty estimates. In addition, our approach yields reliable confidence estimates without the need for verbalization, or external supervision, while preserving full generative expressiveness and structured reasoning behavior.

\section{Methodology}
\subsection{PPO for Language-Based
Quantitative Prediction}
Following prior work, we model language generation as a Markov Decision Process (MDP), where states correspond to token sequences and the initial state, $s_0$, is given by the input prompt~\citep{shoaeinaeini2024guiding, stangel2025rewarding}. At each timestep $t$, the actor (policy) $\pi_\theta$ selects an action $a_t \sim \pi_\theta(\cdot \mid s_t)$,
corresponding to the next generated token. Full details of the MDP formulation are provided in Appendix~\ref{sec:mdp}.

We optimize the policy using an actor–critic framework based on Proximal Policy Optimization (PPO)~\citep{schulman2017proximal}, which maximizes a clipped surrogate objective:
\vspace{0mm}
\begin{equation}\label{eqn:ppo}
    \mathcal{L}_{\text{PPO}}(\theta) =
\mathbb{E}_t \left[
\min \left(
r_t(\theta) A_t,
\text{clip}\left(r_t(\theta), 1 - \epsilon, 1 + \epsilon \right) A_t
\right)
\right],
\end{equation}
where $r_t(\theta) = \frac{\pi_\theta(a_t \mid s_t)}{\pi_{\theta_{\text{old}}}(a_t \mid s_t)}$
denotes the importance sampling ratio between the current policy $\pi_\theta$ and the behavior policy
$\pi_{\theta_{\text{old}}}$ used to collect trajectories, and $\epsilon > 0$ is a clipping hyperparameter that constrains the magnitude of policy updates. 
We compute the advantage estimate $A_t$ in Equation~\ref{eqn:ppo} using Generalized Advantage Estimation (GAE)~\citep{schulman2015high}:
\vspace{-1mm}
\begin{equation}
    {A}_t = \sum_{k=0}^{T-t-1} (\gamma \lambda)^k \, \delta_{t+k},
    \label{eqn:gae}
\end{equation}
where $\lambda \in [0,1]$ controls the bias--variance tradeoff of the estimator, 
$\gamma \in (0,1]$ is the discount factor, 
$\delta_t = R(s_t) + \gamma V_\phi(s_{t+1}) - V_\phi(s_t)$ denotes the temporal-difference error, and $R(s_t)$ is the reward received at state $s_t$.

Here, the critic model $V_\phi$ estimates the expected return from a given state $s_t$ and is trained concurrently with the actor by minimizing the value loss:
\begin{equation}
    \mathcal{L}_{V}(\phi) =
\mathbb{E}_t \left[ \left( V_\phi(s_t) - \hat{R}_t \right)^2 \right],
\label{eqn:value_loss}
\end{equation}
where $\hat{R}_t$ denotes the empirical return. Following standard RLVR setups, we employ a terminal reward structure with reward function $R(s_T) \in \mathbb{R}$, such that rewards are provided only at the terminal state $s_T$. Consequently, the return takes the form $\hat{R}_t = \gamma^{T-t} R(s_T)$. In our setting, the terminal reward is computed based on the model’s final quantitative estimate produced after generating a reasoning trajectory in response to the input query.

Under this formulation, the actor is incentivized to select actions that improve this expected return, while the critic learns to estimate the expected outcome of a partial reasoning trajectory. In the following subsection, we describe our reward design, which aligns the critic's value estimates with the actor's numeric prediction error, enabling it to serve as an error-aligned confidence estimator.

\subsection{CARE: Confidence-Aligned Reward for Estimation}

We design the reward function $R(s_T)$ to serve a dual purpose: \emph{(i)} providing a dense, error-sensitive training signal that captures degrees of correctness and guides the actor toward accurate numerical predictions, and \emph{(ii)} inducing a critic value function that can serve as an error-aligned confidence estimator at inference time.

\paragraph{Reward Definition} 

Let $\hat{y}$ denote the actor's final numerical prediction extracted from the generated trajectory, and let $y$ denote the ground-truth value. We define the absolute prediction error as $\xi = |\hat{y} - y|$ and  the terminal reward as a monotonically decreasing function of prediction error:
\begin{equation}
    R(s_T) = h(\xi),
    \label{eqn:care}
\end{equation}
where $h:\mathbb{R}_{\geq 0} \rightarrow \mathbb{R}$ is chosen so that smaller prediction errors receive larger rewards. This yields an error-sensitive training signal for the actor. Since the critic is trained to estimate returns under this reward, its value function becomes aligned with prediction quality, enabling it to serve as a confidence signal at inference time.

\paragraph{Reward Instantiations} 

We consider two instantiations of the reward function $h$, corresponding to different ways of mapping prediction error to reward. 

\textbf{I: Linear reward:} 
A natural choice is a linear function of prediction error:

\begin{equation}
    h_{\mathrm{linear}}(\xi) = -\xi,
    \label{eqn:care_linear}
\end{equation}

which directly preserves the magnitude of the error and provides a dense, unbounded training signal. While this formulation encourages accurate prediction by penalizing larger errors more strongly, it does not produce a normalized or bounded target for the critic, making its interpretation as a confidence score less direct.

\textbf{II: Gaussian reward.}
To obtain a bounded and interpretable confidence signal, we also consider a scaled Gaussian kernel over the prediction error:
\begin{equation}
    h_{\mathrm{gauss}}(\xi)
    =
    \alpha\exp\left(-\frac{\xi^2}{2\sigma^2}\right),
    \label{eqn:care_gaussian}
\end{equation}
where $\sigma$ controls the task-specific error tolerance and $\alpha > 0$ is a reward scaling factor. This reward is smooth and bounded in $[0, \alpha]$, assigning high reward to small errors and decaying toward zero as error increases. The parameter $\sigma$ has a natural interpretation as a tolerance scale: predictions within $\pm \sigma$ of the ground truth receive high reward, while larger errors are penalized progressively.

We refer to models trained with the linear and Gaussian reward instantiations as \careppo{}-Linear and \careppo{}-Gaussian, respectively.

\section{Experimental Setup}
\subsection{Tasks \& Datasets}
\label{sec:datasets}
We evaluate our approach on two real-world quantitative estimation tasks spanning healthcare and finance, both requiring free-form natural language inputs to produce numerical predictions.

\paragraph{Nutrition Estimation}
We study carbohydrate estimation from natural-language meal descriptions, as illustrated in Figure~\ref{fig:pipeline}, motivated by its importance for diabetes management~\cite{buck2022evaluation}. We generate natural-language meal descriptions from both individual food items and combinations of food items derived from the FoodData Central (FDC) repository~\citep{fdc} using GPT-4o-mini (see Appendix~\ref{sec:appendix_data_nutrition} for details and examples). The resulting dataset contains 26,549 training examples and 6,226 validation examples. For testing, we use the \nutri{} benchmark~\cite{hua2024nutribench}, which consists of 4,826 examples.




\paragraph{Product Price Prediction}
Product price prediction from textual product descriptions is evaluated using the Amazon Reviews dataset~\cite{hou2024bridging}. For our primary experiments, we focus on the Appliances category, where each input is constructed from product metadata fields (e.g., name, category, rating, and descriptive attributes). Full preprocessing details and examples are provided in Appendix~\ref{sec:appendix_data_price}. The resulting dataset contains 16,082 training, 946 validation, and 3,784 test examples.


\subsection{Baselines}
To evaluate the \careppo{} framework and its critic-based confidence estimation, we compare against several training and confidence-estimation baselines.

\paragraph{Training Methods:}
\begin{itemize}[leftmargin=*]
    \item \textbf{Base Model}: The instruction-tuned Qwen-3 4B and 8B models, without task-specific fine-tuning
    \item \textbf{SFT}: Models fine-tuned via next-token prediction on task-specific chain-of-thought data. For nutrition estimation, we construct rule-based reasoning traces that decompose meal descriptions into item-level carbohydrate estimates and aggregate them. For price prediction, we follow prior work on reasoning distillation~\citep{guo2025deepseek, ho2023large} by generating reasoning traces using a stronger teacher model (GPT-4o-mini). Examples of reasoning traces are provided in Appendix~\ref{sec:appendix_data_construction}.
    \item \textbf{PPO-Binary}: A PPO baseline using a binary correctness reward $h_{\mathrm{binary}}(\xi) = \mathbb{I}[\xi \leq \tau]$, where $\tau$  is a task-specific tolerance threshold. This mirrors reinforcement learning from verifiable rewards (RLVR) setups~\citep{wen2025reinforcement} and serves as a direct comparison to evaluate whether error-sensitive reward shaping is necessary for language-based quantitative prediction.
\end{itemize}

\paragraph{Confidence Estimation Baselines}
\begin{itemize}[leftmargin=*]
    \item \textbf{Answer Probability:} We calculate the mean probability of the generated tokens in the prediction string as a measure of model confidence \cite{tian2023just}.
    \item \textbf{EntropyInv:} We derive a confidence metric from the average per-token entropy ($H(s) = -\sum p \log p$) of the generated sequence, a commonly used measure of uncertainty \cite{huang2023look}. We compute entropy over normalized top-20 token probabilities at each decoding step. To ensure comparability, we further normalize the average entropy by the maximum entropy $H_{max}$ and define confidence as the complement: $C_{ent} = 1 - \frac{\bar{H}}{H_{max}}$. We call this method EntropyInv.
    \item \textbf{Verbalized Confidence:} Prior work has shown that verbalized confidence methods can outperform logit-based approaches for reinforcement learning–fine-tuned models~\cite{tian2023just}. We evaluate two methods, (i) \textbf{Verb.~1S (one-stage)}, where the model is prompted to produce both the prediction and a confidence estimate in a single response, and (ii) \textbf{Verb.~2S (two-stage)}, where the model first produces its prediction and is subsequently prompted to assign a confidence score~\cite{tian2023just}.

\end{itemize}

\subsection{Evaluation Metrics}
\paragraph{Quantitative Estimation Performance} We evaluate quantitative prediction accuracy using the mean absolute error (MAE) between model predictions and ground-truth values, where lower values indicate better performance.

\paragraph{Confidence Alignment} We evaluate how well estimated confidence reflects prediction error using the Area Under the Sparsification Error (AUSE)~\citep{ilg2018uncertainty} for assessing the utility of confidence estimates for selective prediction. AUSE is computed as the area between a sparsification curve, obtained by progressively removing predictions in order of lowest confidence and measuring the error of the remaining set, and an oracle curve obtained by removing predictions in order of highest true error. A lower AUSE indicates that confidence more effectively ranks predictions by reliability. 


\subsection{Training Details}
Both actor and critic are initialized from instruction-tuned Qwen-3 models (4B and 8B). We generate 8 rollouts per prompt and use GAE with $\gamma=\lambda=1.0$. Actor and critic learning rates are set to $10^{-6}$ and $10^{-5}$, respectively. Models are trained for 5 epochs using VERL~\cite{verl} on a single 8-GPU node.

For SFT, we select the checkpoint with the lowest validation MAE. For PPO, we select the actor checkpoint with the lowest validation MAE and the corresponding critic with the best AUSE.

For the Gaussian reward formulation $h_{\mathrm{gauss}}$, we set $\sigma = 10$, aligning with the error scale in carbohydrate estimation, where prior work considers errors within approximately $7.5$g to be a conservative acceptable range~\citep{hua2024nutribench}. Under this parameterization, moderate errors (e.g., $\xi \approx 7.5$g) correspond to relatively high confidence ($\approx 0.75$), while large errors (e.g., $\xi \approx 30$g) map to near-zero confidence. We apply the same $\sigma$ for price prediction, as both tasks operate over comparable output ranges in our datasets.
We evaluate sensitivity to moderate variations in $\sigma$ by conducting an ablation with $\sigma \in \{8, 9, 10, 11, 12\}$ using Qwen3-4B on the nutrition estimation task (Section~\ref{sec:sensitivity_sigma}), and find that performance remains relatively stable across nearby values, particularly for confidence alignment. We use a reward scaling factor of $\alpha=8$. Since the confidence score is obtained by normalizing the critic value by $\alpha$ (Equation~\ref{eqn:care_gaussian}), this choice does not affect the resulting confidence scale.

We set the correctness threshold for $h_{\mathrm{binary}}$ to $7.5$g for carbohydrate estimation~\citep{hua2024nutribench} and 5 USD for price prediction, reflecting moderate task-specific error tolerances. When no extractable prediction is produced, we assign the minimum reward value for each formulation: $0$ for the bounded rewards $h_{\mathrm{gauss}}$ and $h_{\mathrm{binary}}$, and $-500$ for the unbounded reward $h_{\mathrm{linear}}$, chosen based on the upper range of prediction errors observed on the base model validation sets.

\section{Results}

\subsection{CARE-PPO Jointly Improves Quantitative Accuracy and Confidence Alignment}
\label{sec:res_indomain}

\begin{figure}[t]
    \centering
    \includegraphics[width=\linewidth]{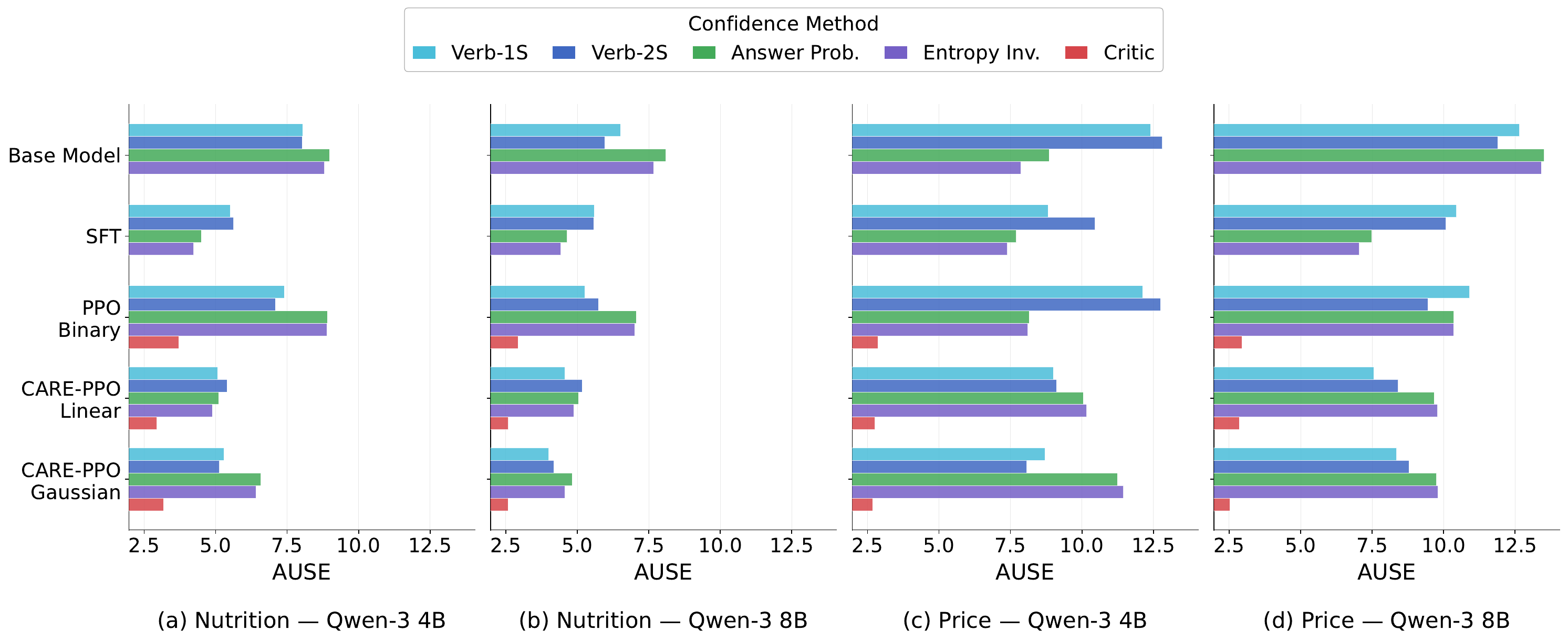}
    \caption{Confidence alignment (AUSE) across confidence estimation methods, training approaches, tasks, and model scales. Lower values indicate better confidence alignment. Across all settings, critic-based confidence estimation yields the strongest overall alignment. Corresponding sparsification curves used to compute AUSE are visualized in Appendix~\ref{sec:full_numerical_results} (Figure~\ref{fig:sparsi_indist}).}
    \label{fig:confidence_methods}
\end{figure}

Table~\ref{tab:indomain} reports quantitative prediction performance (MAE) across all training methods, tasks, and model scales. Figure~\ref{fig:confidence_methods} shows confidence alignment results (AUSE) for each confidence estimation method paired with each training approach. Since the Verb-1S method uses a different prompt that generates a new prediction alongside the confidence estimate, its MAE may differ slightly; we report the lowest MAE per method, with full results provided in Appendix~\ref{sec:appendix_indomain}.

Across both tasks and model scales, both reward variants of \careppo{} (\careppo{}-Linear and \careppo{}-Gaussian) substantially reduce MAE relative to the base model and consistently outperform PPO-Binary. This improvement over PPO-Binary suggests that continuous, error-sensitive rewards provide a more effective training signal than binary correctness rewards, which discard error magnitude information. While SFT achieves lower MAE, it comes at the cost of poor confidence alignment: SFT's confidence estimates show weak alignment with prediction error across both tasks, regardless of the confidence estimation method used.

\begin{table*}[t]
\centering
\caption{Quantitative estimation performance (MAE) on nutrition (carbohydrate) estimation and product (appliances) price prediction. We report the lowest MAE per training method across confidence estimators (see Appendix~\ref{sec:appendix_indomain} for full results). Lower is better ($\downarrow$).}
\label{tab:indomain}
\renewcommand{\arraystretch}{1.15}
\setlength{\tabcolsep}{5pt}
\small
\begin{tabular}{l cc cc}
\toprule
\multirow{2}{*}{\textbf{Training Method}}
& \multicolumn{2}{c}{\textbf{Nutrition Estimation MAE $\downarrow$}}
& \multicolumn{2}{c}{\textbf{Price Prediction MAE $\downarrow$}} \\
\cmidrule(lr){2-3} \cmidrule(lr){4-5}
& \textbf{Qwen-3 4B} & \textbf{Qwen-3 8B}
& \textbf{Qwen-3 4B} & \textbf{Qwen-3 8B} \\
\midrule
Base Model        & 18.07 & 15.37 & 18.77 & 18.62 \\
SFT               & \textbf{10.23} & \textbf{9.51}
                  & \textbf{12.48} & \textbf{11.59} \\
PPO-Binary        & 14.58 & 12.38 & 14.97 & 15.03 \\

\rowcolor{careblue}
CARE-PPO-Linear   & \underline{11.07} & 11.43
                  & \underline{12.56} & \underline{12.69} \\

\rowcolor{careblue}
CARE-PPO-Gaussian & 11.50 & \underline{10.42}
                  & 14.37 & 13.57 \\
\bottomrule
\end{tabular}
\end{table*}

In contrast, as shown in Figure~\ref{fig:confidence_methods}, \careppo{} consistently produces significantly better-aligned confidence estimates, with the critic emerging as the strongest confidence estimator across all methods, models, and settings. We visualize the corresponding sparsification curves used to compute AUSE in Appendix~\ref{sec:full_numerical_results} (Figure~\ref{fig:sparsi_indist}). Even for PPO-Binary, where the reward provides no information about error magnitude, the critic achieves lower AUSE than all verbalized and logit-based baselines, supporting our hypothesis that the actor--critic training dynamic naturally encourages the critic to learn a meaningful confidence signal even from a discretized reward. When the reward is directly aligned with prediction error, as in \careppo{}-Linear and \careppo{}-Gaussian, this effect is further amplified, suggesting that closer alignment between reward and prediction loss leads to more reliable confidence estimation. 
Section~\ref{sec:critic_dynamics} additionally visualizes how critic output distributions evolve throughout training and progressively align with the underlying reward structure.

\subsection{CARE-PPO is Robust in Out-of-Distribution Settings}
\label{sec:res_ood}

We evaluate \careppo{} under two real-world out-of-distribution settings: linguistic shift in nutrition estimation and domain shift in product price prediction.

For nutrition estimation, we test on multilingual meal descriptions, reflecting realistic scenarios where users describe foods in different languages with culturally specific terms and dietary conventions~\citep{winata2025worldcuisines}. We construct this test set from WHO dietary records~\citep{fao_who_gift} spanning Spanish, Italian, Portuguese, Tagalog, and Amharic, using GPT-4o-mini to generate natural-language meal descriptions from the food records (Appendix~\ref{sec:appendix_data_nutrition}). We retain paired local-language and English descriptions from the source records rather than relying on automated translation, preserving official terminology and reducing translation-induced artifacts.

For price prediction, users may query about products across a broad range of categories. Thus, to assess generalization under domain shift, we additionally use the Electronics category from the Amazon Reviews dataset~\cite{hou2024bridging} as an out-of-distribution test set, randomly sampling 4,000 examples.

\begin{table}[!t]
\centering
\caption{Quantitative estimation performance (MAE) under linguistic shift for nutrition estimation (English-trained models evaluated on multilingual queries) and domain shift for price prediction (Appliances-trained models evaluated on Electronics). We report the lowest MAE per training method across confidence estimators (see Appendix~\ref{sec:appendix_ood} for full results). Lower is better ($\downarrow$).}
\label{tab:ood_results}
\renewcommand{\arraystretch}{1.15}
\setlength{\tabcolsep}{5pt}
\small
\begin{tabular}{l cc cc}
\toprule

\multirow{2}{*}{\textbf{Training Method}}
& \multicolumn{2}{c}{\begin{tabular}{@{}c@{}}\textbf{Nutrition Estimation MAE ($\downarrow$)}\\[-1pt]{\small \emph{Linguistic Shift}}\end{tabular}}
& \multicolumn{2}{c}{\begin{tabular}{@{}c@{}}\textbf{Price Prediction MAE ($\downarrow$)}\\[-1pt]{\small \emph{Domain Shift}}\end{tabular}} \\

\cmidrule(lr){2-3} \cmidrule(lr){4-5}

& \textbf{Qwen-3 4B} & \textbf{Qwen-3 8B}
& \textbf{Qwen-3 4B} & \textbf{Qwen-3 8B} \\

\midrule

Base Model
& 32.47 & 28.15
& 29.13 & 23.36 \\

SFT
& 30.02 & 24.95
& \textbf{17.56} & 16.88 \\

PPO-Binary
& 26.06 & \underline{23.36}
& \underline{19.02} & 17.25 \\

\rowcolor{careblue}
CARE-PPO-Linear
& \textbf{23.48} & \textbf{20.64}
& 20.47 & 17.01 \\

\rowcolor{careblue}
CARE-PPO-Gaussian
& \underline{24.00} & 24.26
& 19.17 & \textbf{15.79} \\

\bottomrule
\end{tabular}
\end{table}
\begin{figure}[t]
    \centering
    \includegraphics[width=\linewidth]{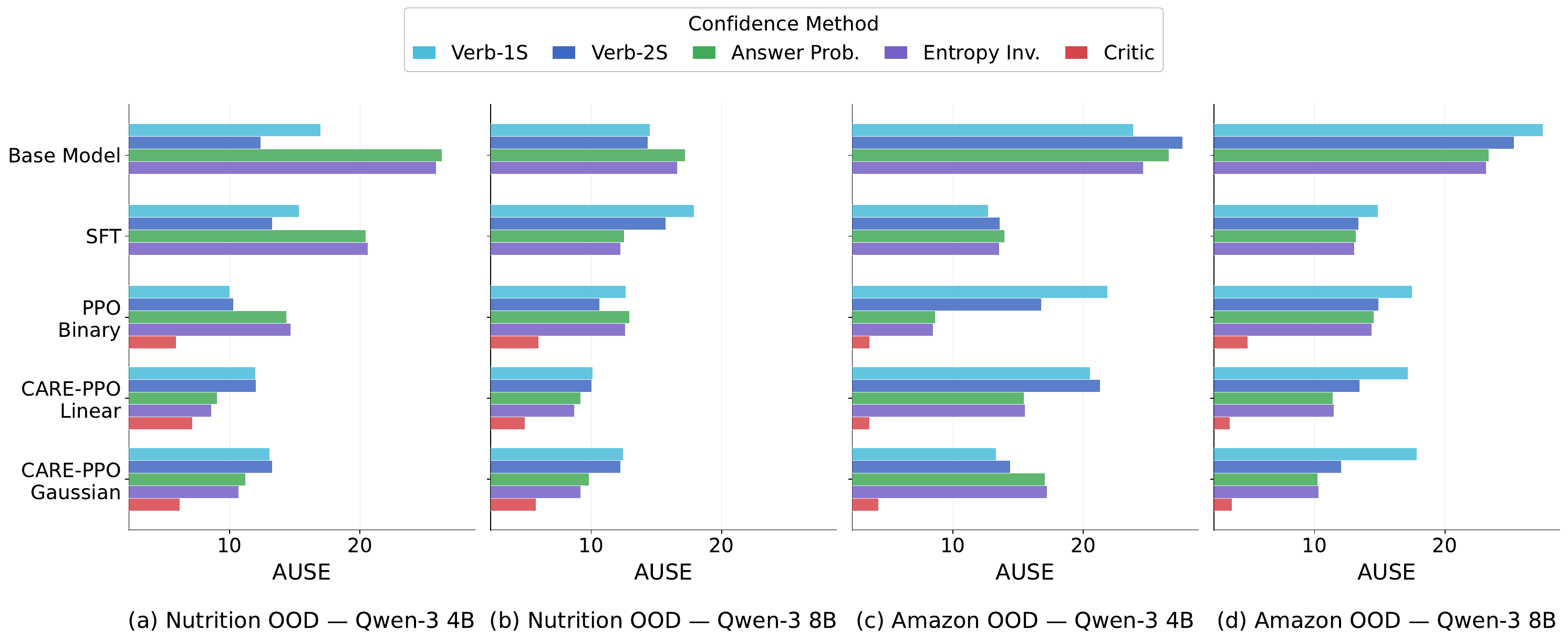}
    \caption{Confidence alignment (AUSE) across confidence estimation methods, training approaches, tasks, and model scales in out-of-distribution settings. Lower values indicate better confidence alignment. Across all settings, critic-based confidence estimation yields the strongest overall alignment. Corresponding sparsification curves used to compute AUSE are visualized in Appendix~\ref{sec:full_numerical_results} (Figure~\ref{fig:sparsi_ood}).}
    \label{fig:confidence_methods_ood}
\end{figure}

\subsubsection{Linguistic Shift in Nutrition Estimation}

Table~\ref{tab:ood_results} (left) reports MAE under linguistic shift, where models trained on English meal descriptions are evaluated on multilingual queries spanning five languages. Notably, \careppo{} achieves stronger MAE performance than SFT under multilingual shift, suggesting that PPO-based approaches may learn representations that generalize more robustly across languages. Additional analysis in Table~\ref{tab:delta_mae} in the Appendix further shows that PPO-based methods exhibit smaller MAE gaps between parallel English and non-English queries describing the same meal than SFT and the base model. Simultaneously, Figures~\ref{fig:confidence_methods_ood} \textit{(a)} and \textit{(b)} demonstrate that critic-based confidence estimates consistently achieve substantially lower AUSE than verbalized and logit-based confidence estimation methods across both model scales, demonstrating strong robustness under linguistic shift.



\subsubsection{Domain Shift in Price Prediction}

Table~\ref{tab:ood_results} (right) reports MAE, while Figures~\ref{fig:confidence_methods_ood} \textit{(c)} and \textit{(d)} show the AUSE of confidence estimates under domain shift, where models trained on Appliances are evaluated on unseen Electronics products. Confidence alignment degrades substantially for the base model and SFT, whereas critic-based confidence estimates learned through PPO (both PPO-Binary and the \careppo{} variants) remain substantially better aligned under shift while maintaining competitive MAE performance.

Overall, across both linguistic and domain shifts, critic-based confidence estimates learned through PPO remain better aligned, more stable, and more robust than verbalized or logit-based alternatives, suggesting that such confidence signals are more likely to remain reliable under the distribution shifts encountered in practical deployment settings.

\subsubsection{CARE-PPO Preserves Task-Aware Behavior}
\label{sec:forgetting}

In practice, users interact with LLMs in open-ended settings and may issue queries beyond the deployment use case, making it important to preserve general-purpose behavior even after specialization to a quantitative prediction task. For instance, a model fine-tuned for nutrition estimation should recognize when a query is unrelated to food and respond appropriately, rather than attempting to estimate carbohydrates. We therefore evaluate whether CARE-PPO exhibits more task-aware behavior on general-purpose queries outside the training distribution, specifically studying whether models incorrectly force task-specific predictions on unrelated inputs.

\begin{table}[t]
\centering
\caption{Task overfitting evaluation for nutrition estimation on 500 general instruction-following queries from IFEval~\citep{zhou2023instruction}. We report the proportion of responses that incorrectly treat unrelated inputs as meal descriptions and produce carbohydrate estimates. Lower is better ($\downarrow$).}
\label{tab:task_aware}
\small
\setlength{\tabcolsep}{8pt}
\renewcommand{\arraystretch}{1.1}
\begin{tabular}{lccccc}
\toprule
\makecell{\textbf{Model}\\\textbf{Backbone}}
& \makecell{\textbf{Base}\\\textbf{Model}}
& \textbf{SFT}
& \makecell{\textbf{PPO}\\\textbf{Binary}}
& \cellcolor{careblue} \makecell{\textbf{CARE-PPO}\\\textbf{Linear}}
& \cellcolor{careblue} \makecell{\textbf{CARE-PPO}\\\textbf{Gaussian}} \\
\midrule
Qwen-3 4B & 33\% & 73\% & \textbf{1\%} & \cellcolor{careblue} 29.2\% & \cellcolor{careblue} \underline{7.6\%} \\
Qwen-3 8B & 41.8\% & 33.2\% & \underline{12\%} & \cellcolor{careblue} 16.8\% & \cellcolor{careblue} \textbf{11.2\%} \\
\bottomrule
\end{tabular}
\end{table}



To evaluate this behavior, we test models on 500 randomly sampled queries from the instruction-following benchmark IFEval~\citep{zhou2023instruction}, which contains prompts unrelated to the target prediction tasks. We focus on nutrition estimation as it provides the clearest signal of task overfitting: interpreting an unrelated query as a meal description and producing a carbohydrate estimate is an unambiguous failure mode. In contrast, for price-trained models, certain open-ended prompts (e.g., itinerary planning or writing assistance) can plausibly be interpreted as requests for a service to be priced, making it harder to distinguish genuine overfitting from reasonable model behavior. We use GPT-4o-mini as an LLM-as-a-judge (LLMaJ)~\citep{gu2024survey} to classify each response based on whether the model incorrectly produces a task-specific prediction or responds appropriately. We provide the LLMaJ prompt in Appendix~\ref{sec:llmaj_prompt}.


Table~\ref{tab:task_aware} shows that SFT exhibits substantial task overfitting, frequently misinterpreting unrelated queries as meal descriptions. Representative qualitative examples of these behaviors are provided in Appendix~\ref{sec:appendix_task_aware_examples}. In contrast, all PPO-based methods substantially reduce this behavior, consistent with prior findings that RL fine-tuning generalizes more robustly beyond the training distribution than SFT~\citep{chu2025sft}. Among PPO variants, PPO-Binary and CARE-PPO-Gaussian exhibit the lowest misinterpretation rates, while CARE-PPO-Linear exhibits notably higher rates. We hypothesize that this difference is partly attributable to reward design: because $h_{\text{linear}}$ is unbounded, the penalty assigned when no extractable prediction is produced is significantly larger in magnitude than for the bounded formulations $h_{\text{gauss}}$ and $h_{\text{binary}}$ , potentially incentivizing the actor to always produce a formatted numerical output regardless of input relevance. Overall, these results suggest that PPO-based fine-tuning more effectively preserves general-purpose behavior beyond the target prediction setting, indicating better robustness in open-ended deployment settings.

\subsection{Critic Distribution Evolution During Training}
\label{sec:critic_dynamics}

\begin{figure*}[!h]
    \centering
    \includegraphics[width=\linewidth]{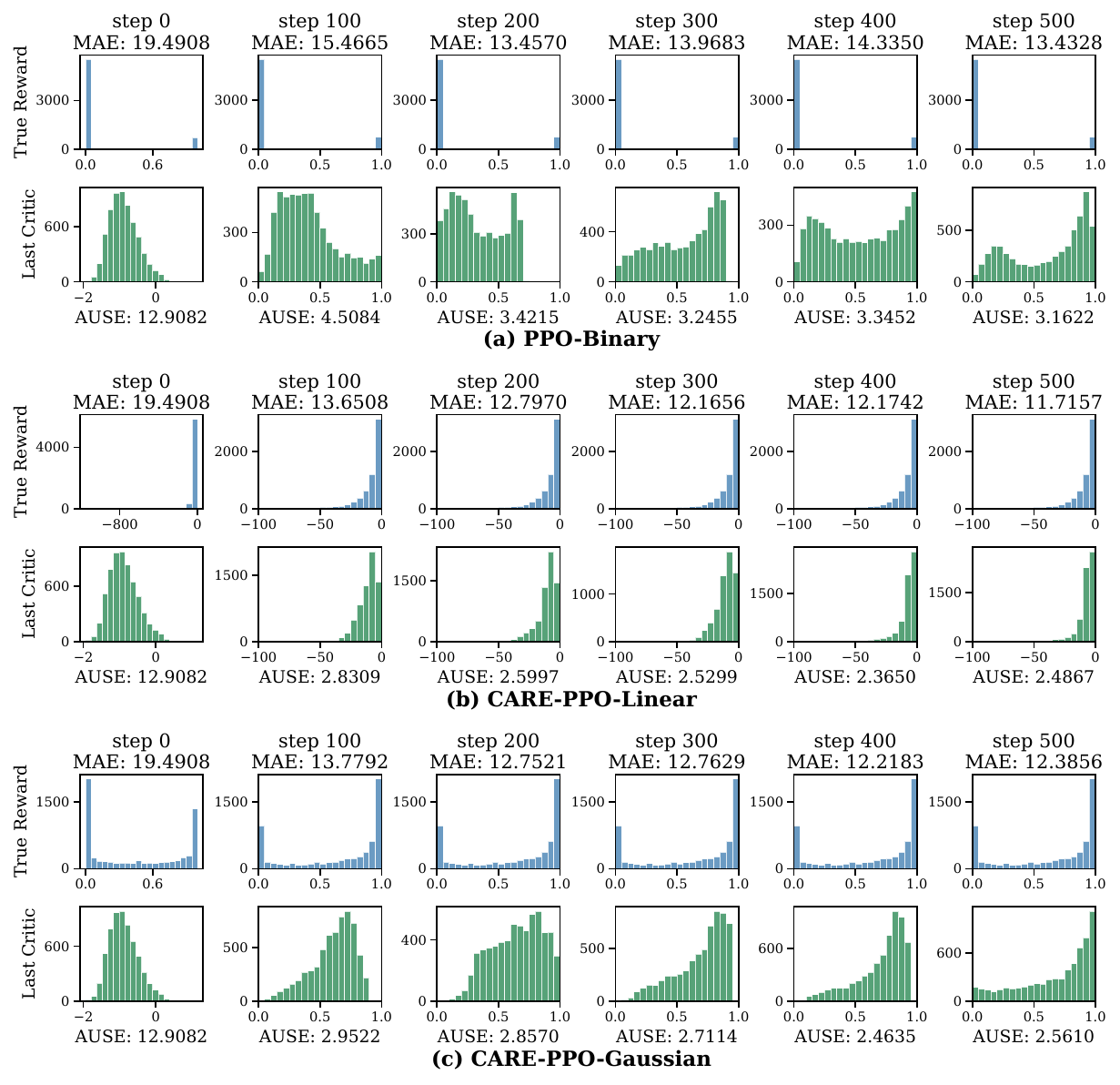}
    \caption{Evolution of critic output distributions throughout PPO training on the validation set for PPO-Binary, \careppo{}-Linear, and \careppo{}-Gaussian for the nutrition estimation task with Qwen-3 4B as the backbone model. For each checkpoint, we visualize the distribution of ground-truth rewards (top) and the corresponding final-layer critic outputs (bottom). As training progresses, critic outputs increasingly reflect the structure of the reward distribution, corresponding to improved confidence alignment and lower AUSE.}
    \label{fig:critic_dynamics}
\end{figure*}

To better understand how critic-based confidence estimation emerges during training, we visualize the distribution of critic outputs throughout PPO optimization alongside the corresponding ground-truth reward distributions on the validation set. Figure~\ref{fig:critic_dynamics} shows the evolution of the final-layer critic outputs across training checkpoints for PPO-Binary, \careppo{}-Linear, and \careppo{}-Gaussian trained for the nutrition estimation task with Qwen-3 4B as the backbone. Corresponding plots for the price prediction task are presented in Appendix~\ref{sec:critic_dist_price}.

Across all reward formulations, the critic progressively learns to produce confidence estimates that better reflect the structure of the underlying reward distribution. In particular, the critic distributions become increasingly shaped according to prediction quality over training, corresponding to improved confidence alignment. While PPO-Binary operates with a discretized reward signal, the critic nevertheless learns a meaningful ranking over prediction quality, supporting our hypothesis that actor--critic optimization naturally encourages confidence-aware representations even under sparse reward formulations.

\subsection{Sensitivity Analysis with $\sigma$ on \careppo{}-Gaussian}
\label{sec:sensitivity_sigma}

\begin{wraptable}{r}{0.42\textwidth}
\centering
\vspace{-1.2em}
\caption{Sensitivity analysis of the Gaussian reward tolerance parameter $\sigma$. Lower is better ($\downarrow$) for both MAE and AUSE.}
\label{tab:sigma_ablation}
\small
\setlength{\tabcolsep}{15pt}
\renewcommand{\arraystretch}{1}
\begin{tabular}{ccc}
\toprule
$\sigma$ & MAE $\downarrow$ & AUSE $\downarrow$ \\
\midrule
8  & 15.92 & 3.70 \\
9  & 12.13 & 3.06 \\
10 & 11.72 & 3.18 \\
11 & 13.89 & 3.39 \\
12 & 12.19 & 3.10 \\
\bottomrule
\end{tabular}
\vspace{-1em}
\end{wraptable}

The Gaussian reward formulation in \careppo{}-Gaussian introduces a tolerance parameter $\sigma$ that controls the smoothness of the reward landscape around the target value. To evaluate sensitivity to this hyperparameter, we conduct an ablation over $\sigma \in \{8, 9, 10, 11, 12\}$ using Qwen-3 4B on the nutrition estimation task. Results are shown in Table~\ref{tab:sigma_ablation}.

Overall, we find that performance remains relatively stable across nearby values of $\sigma$, particularly in terms of confidence alignment. While $\sigma=10$ achieves the best overall balance between quantitative prediction accuracy and confidence calibration, neighboring values produce comparable AUSE performance, suggesting that \careppo{}-Gaussian is reasonably robust to moderate variations in the reward tolerance parameter.


\section{Conclusion}
We introduced \careppo{}, a reinforcement learning framework for language-based quantitative prediction that draws a novel connection between loss prediction for uncertainty estimation and actor--critic RL fine-tuning. By defining rewards as functions of prediction error, \careppo{} enables the critic to learn value functions aligned with prediction quality, allowing the critic to serve as an effective confidence estimator at inference time. Across quantitative prediction tasks in healthcare and finance, \careppo{} achieves competitive prediction accuracy while substantially improving confidence alignment relative to verbalized and logit-based confidence estimation methods. We further showed that these gains remain robust under realistic distribution shifts and extend to improved task-aware behavior on out-of-task queries. 

We believe confidence-aware language-based quantitative prediction represents a promising research direction. While \careppo{} relies on PPO fine-tuning, confidence estimation emerges as a byproduct of actor--critic training rather than requiring a separate confidence-learning stage, calibration procedure, or auxiliary model. Latency analysis shows that critic-based confidence estimation introduces only a small inference overhead (Appendix~\ref{sec:appendix_latency}), suggesting that it can be incorporated into deployment with minimal additional cost. Our experiments are currently limited to models up to 8B parameters and two application domains due to computational constraints. Future work will explore scaling these results to larger language models and broader quantitative prediction settings, investigating theoretical connections between reward design and calibration, and integrating confidence estimates more directly into the generation process to more tightly couple prediction and uncertainty estimation.

\bibliography{ref}
\bibliographystyle{unsrt}







\appendix
\section{PPO Framework}
\label{sec:mdp}
We follow prior work in modeling language generation as a Markov Decision Process (MDP) defined by the tuple
$\left(\mathcal{S}, \mathcal{A}, \mathcal{P}, R, \gamma\right)$~\citep{shoaeinaeini2024guiding, stangel2025rewarding}.
Each state $s_t \in \mathcal{S}$ represents a sequence of tokens drawn from a fixed vocabulary $\mathcal{V}$.
The initial state $s_0 = (x_1, x_2, \dots, x_L)$, where $x_i \in \mathcal{V}$, corresponds to the input prompt,
including the system instruction and user query.
At each timestep $t \in \{0, \dots, T-1\}$, the actor (policy) model $\pi_\theta$ selects an action
$a_t \sim \pi_\theta(\cdot \mid s_t)$, where $a_t \in \mathcal{V}$ corresponds to generating the next token.
The environment transition function $\mathcal{P}$ is deterministic, forming the next state by appending the selected action to the
current sequence, $s_{t+1} = (s_t, a_t)$.
Generation terminates at a terminal state $s_T$ upon emitting an end-of-sequence token or reaching a maximum sequence length $T$.
The discount factor $\gamma \in (0,1]$ controls the contribution of the terminal reward to earlier states.

\section{Data Construction and Processing}
\label{sec:appendix_data_construction}

\subsection{Nutrition Estimation}
\label{sec:appendix_data_nutrition}
We construct training data for the nutrition estimation task using food composition entries from FoodData Central (FDC)~\cite{fdc}, the food composition database maintained by the United States Department of Agriculture (USDA). Specifically, we first obtain food names along with their corresponding nutritional content for varying portion sizes.

To convert these structured food entries into natural-language training examples, we prompt GPT-4o-mini to generate realistic free-form meal descriptions from the food items. Concretely, we provide the model with food names and portion sizes from FDC entries and instruct it to synthesize meal descriptions in natural language.

In addition to constructing descriptions directly from individual food entries, we also generate meal descriptions containing combinations of two food items to better reflect realistic meals consisting of multiple components. The prompts used for meal description generation are shown in Box~\ref{box:nutrition-generation-prompts}.

\begin{promptbox}[label={box:nutrition-generation-prompts}]
{Nutrition Estimation Data Generation Prompts}

\textit{\textbf{Single-Item Meal Description Generation Prompt}}

\vspace{0.5em}

\textbf{Role:} You are a meal description bot. Your style should be conversational and natural. Your tone should be matter-of-fact. Do not use flowery or overly descriptive words or phrasing.

\vspace{0.5em}

\textbf{Objective:} Given an input, describe the meal in plain language as a regular person would.

\vspace{0.5em}

\textbf{Instructions:}
\begin{itemize}
    \item Provide five different meal descriptions in a dictionary with keys ``description1'', ``description2'', etc.
    \item Extract the \textbf{exact food names and serving sizes} from each description and store them in the ``items'' field.
    \item Ensure consistency between ``description'' and ``items''---the ``items'' field must not omit details that appear in the description.
    \item Return only the meal descriptions and items without additional information.
\end{itemize}

\vspace{0.5em}

\textbf{Context:} Use the following examples of user input and model output as a guide:

\vspace{0.5em}

\textbf{Input1}
\begin{verbatim}
{
  "parts": [
    "smoothie, made with spinach, banana, almond milk, protein powder"
  ],
  "units": [
    "1.0 glass"
  ]
}
\end{verbatim}

\textbf{Output1}
\begin{verbatim}
{
  "description1": {
    "description": "I am having a glass of smoothie made with spinach,
                    banana, almond milk, and protein powder for lunch.",
    "items": "a glass of smoothie made with spinach, banana,
              almond milk, and protein powder"
  },
  etc
}
\end{verbatim}

\textbf{Input2}
\begin{verbatim}
{
  "parts": [
    "Beverages, NESTEA, tea, black, ready-to-drink, lemon"
  ],
  "units": [
    "1.0 serving (8 fl oz)"
  ]
}
\end{verbatim}

\textbf{Output2}
\begin{verbatim}
{
  "description1": {
    "description": "8 fl oz of NESTEA lemon black tea",
    "items1": "8 fl oz of NESTEA lemon black tea"
  },
  etc.
}
\end{verbatim}

\textbf{Input3}
\begin{verbatim}
{
  "parts": [
    "pepperoni pizza"
  ],
  "units": [
    "1.0 slice"
  ]
}
\end{verbatim}

\textbf{Output3}
\begin{verbatim}
{
  “description1”: {
    "description": "One slice of pepperoni pizza",
    "items1": "one slice of pepperoni pizza"
  },
  etc.
}
\end{verbatim}

\textbf{Input4}
\begin{verbatim}
{
  "parts": [
    "Breakfast tart, lowfat"
  ],
  "units": [
    "1 Pop Tart"
  ]
}
\end{verbatim}

\textbf{Output4}
\begin{verbatim}
{
  “description1”: {
    "description": "I have a low-fat Breakfast Tart.",
    "items1": "a low-fat Breakfast Tart."
  },
  etc.
}
\end{verbatim}

\vspace{0.5em}

\textbf{Task:} For each input, create five different meal descriptions in plain, conversational language. Additionally, output the natural food names including natural serving sizes in the descriptions.

\vspace{1em}

\textit{\textbf{Multi-Item Meal Description Generation Prompt}}

\vspace{0.5em}

\textbf{Task:} Combine two meal descriptions into one, creating a natural-sounding sentence as if spoken by a person. If two queries describe different meals, merge them into a single meal description.

\vspace{0.5em}

\textbf{Input1}
\begin{verbatim}
[
  '43g garlic bread.',
  '100g grilled chicken with steamed vegetables.',
  '100g tomato soup.'
]
\end{verbatim}

\textbf{Output1}
\begin{verbatim}
43g garlic bread, 100g grilled chicken with steamed vegetables, and
100g tomato soup.
\end{verbatim}

\textbf{Input2}
\begin{verbatim}
[
  'For breakfast, I am having 136g banana.',
  'My lunch consists of 106g refrigerated whole wheat naan bread.'
]
\end{verbatim}

\textbf{Output2}
\begin{verbatim}
For lunch, I am having 136g banana and 106g refrigerated whole wheat
naan bread.
\end{verbatim}

\vspace{0.5em}

\textbf{Instruction:} Return only the combined meal descriptions, excluding any additional information.

\vspace{0.5em}

\end{promptbox}

Finally, we generate synthetic reasoning traces for supervised fine-tuning with chain-of-thought supervision. These reasoning traces are constructed using a rule-based procedure that (i) identifies individual food items and their portion sizes, (ii) estimates the carbohydrate content of each item, and (iii) aggregates these values to produce the final carbohydrate estimate. Representative examples of the synthetic reasoning traces are provided in Example~\ref{box:nutrition-examples}.

\begin{examplebox}[label={box:nutrition-examples}]{Nutrition Estimation Training Data Examples}

\small

\textbf{Example 1}

\vspace{0.5em}

\textbf{Query}

100 grams of gluten-free granola with almond butter.

\vspace{0.5em}

\textbf{Synthetic Reasoning Trace}

The meal consists of 100g almond butter gluten-free granola.\newline
100g almond butter gluten-free granola has 60.00g carbs.\newline
So the total carbs in the meal = 60.00.\newline
Output: \{"total\_carbohydrates": 60.00\}

\vspace{1em}

\hrule

\vspace{1em}

\textbf{Example 2}

\vspace{0.5em}

\textbf{Query}

For dinner, I am enjoying a cup of chili with beans that was heated in the microwave.

\vspace{0.5em}

\textbf{Synthetic Reasoning Trace}

The meal consists of a cup of microwavable chili with beans.\newline
A cup of microwavable chili with beans has 26.55g carbs.\newline
So the total carbs in the meal = 26.55.\newline
Output: \{"total\_carbohydrates": 26.55\}

\vspace{1em}

\hrule

\vspace{1em}

\textbf{Example 3}

\vspace{0.5em}

\textbf{Query}

A serving of chunky soup featuring grilled chicken and sausage, weighing 100 grams, and 453.6 grams of trimmed raw leg lamb, focusing on the sirloin half with a quarter inch of fat.

\vspace{0.5em}

\textbf{Synthetic Reasoning Trace}

The meal consists of 100 grams of Grilled Chicken \& Sausage Gumbo Chunky Soup, 453.6 grams of raw leg lamb, sirloin half, trimmed to 1/4 inch of fat.\newline
100 grams of grilled chicken \& sausage gumbo chunky soup has 6.94g carbs.\newline
453.6 grams of raw leg lamb, sirloin half, trimmed to 1/4 inch of fat has 0.00g carbs.\newline
So the total grams of carbs in the meal = (6.94 + 0.00) = 6.94.\newline
Output: \{"total\_carbohydrates": 6.94\}

\end{examplebox}

Further, for out-of-distribution evaluation under linguistic shift, we construct a multilingual nutrition estimation test set using dietary records from the WHO Global Individual Food consumption data Tool (GIFT)~\citep{fao_who_gift}, spanning Spanish, Italian, Portuguese, Tagalog, and Amharic. We use GPT-4o-mini to generate natural-language meal descriptions from the structured dietary records while preserving the original language of each record. The prompt used for multilingual meal description generation is provided in Box~\ref{box:multilingual-nutrition-prompt}. For the final test set, we randomly sample 100 queries from each language. We provide representative meal examples in each language in Example~\ref{box:multilingual-meal-examples}.

\begin{promptbox}[label={box:multilingual-nutrition-prompt}]{Multilingual Nutrition Estimation Data Generation Prompt}
You are generating realistic meal log descriptions for nutrition estimation.

Given food items and weights, write 3 diverse meal descriptions in the target language, like how different real people would log the same meal.

Requirements:
\begin{itemize}
    \item All 3 descriptions must faithfully preserve the food items and quantities. Do not add or infer any nutritional values (calories, macros, etc.) that are not part of the food item names.
    
    \item Vary meaningfully across the following dimensions --- each description should feel distinct:
    \begin{enumerate}
        \item Brevity: one ultra-short (telegraphic), one mid-length, one fuller sentence
        \item Wording: different vocabulary, avoid repeating the same phrasing across descriptions
        \item Sentence structure: mix styles (e.g., bare list, full sentence, diary-style prose)
        \item Quantity explicitness: at least one omits exact grams, at least one includes them
        \item Level of detail: at least one summarizes the meal, at least one names every item
    \end{enumerate}
    
    \item Do not add opinions, subjective commentary (e.g., ``quite filling''), taste/texture descriptions, or cooking inferences unless explicitly stated in the food item name.
    
    \item Do not prefix descriptions with numbers, bullets, or labels.
    
    \item The output should resemble real-world food diary or meal log entries.
    
    \item Some food item names may contain bilingual forms such as ``Crisp eggplant / Mak Keua'' or ``Rice ({\thaifont ข้าว})''. Always use only the name that matches the target language.
    
    \item For non-English outputs, vary sentence structure as much as in English --- avoid repeating the same ``[quantity][food]'' pattern across all 3 descriptions.
\end{itemize}

\end{promptbox}

\begin{examplebox}[label={box:multilingual-meal-examples}]
{Examples of Multilingual Meal Descriptions}

\textbf{Spanish}

\vspace{0.5em}

\textit{Example 1}: Un poco de pan dulce de panadería, unos 70 gramos aproximadamente. (39.47g carbohydrates)

\vspace{0.5em}

\textit{Example 2}: Para la comida, tomé un jugo de naranja industrializado, pesando 156 gramos, acompañado de 242 gramos de sopa de pasta tipo Maruchan. (39.27g carbohydrates)

\vspace{1em}

\textbf{Italian}

\vspace{0.5em}

\textit{Example 1}: Per il bimbo, ho dato 150 grammi di succo di pera Nipiol e 120 grammi di yogurt alla frutta Plasmon con mela e banana. (45.66g carbohydrates)

\vspace{0.5em}

\textit{Example 2}: Per merenda ho mangiato una barretta fitness Nestlé al cioccolato di 27 grammi. (21.82g carbohydrates)

\vspace{1em}

\textbf{Portuguese}

\vspace{0.5em}

\textit{Example 1}: 50g de café, 36g de cuscuz e 100g de ovos mexidos. (12.02g carbohydrates)

\vspace{0.5em}

\textit{Example 2}: Pão integral 50g com margarina 10g e adoçante artificial 2g, acompanhado de duas vitaminas de 300g cada, uma de maçã e outra mista. (119.01g carbohydrates)

\vspace{1em}

\textbf{Tagalog}

\vspace{0.5em}

\textit{Example 1}: Isang maliit na ulam na may 3g gabi, 2g boiled galunggong, 60g white rice, plus 10g bawat isa ng nilutong kalabasa at upo. (20.0g carbohydrates)

\vspace{0.5em}

\textit{Example 2}: Niluto ko ang 40g na gabi, 40g na kalabasa, at 10g na malunggay, sinamahan ng 225g na puting kanin at 10g na tsabita na pinirito. (81.81g carbohydrates)

\vspace{1em}

\textbf{Amharic}

\vspace{0.5em}

\textit{Example 1}: {\amharicfont እኔ ለቁርስ የቦኮሎ ኩፊታ 138 ግራም እና ሙሉ ወተት 197 ግራም ጨምራለሁ።} (47.6g carbohydrates)

\vspace{0.5em}

\textit{Example 2}: {\amharicfont ዛሬ የተመረጡት ምግቦች ይህን ያካተቱ፡ የተለያዩ ከ64 ግራም ቆርን ብርድ (ቅታ) እና 73 ግራም የተከፈለ ምስር ስቲው ያሉበት ምሳ ነበር።} (48.6g carbohydrates)

\end{examplebox}

\subsection{Product Price Prediction}
\label{sec:appendix_data_price}

For the product price prediction task, we use product metadata from the Amazon Reviews dataset~\cite{hou2024bridging}. For our primary experiments, we focus on the Appliances category, which consists of 94.3K items. For each product, we construct a textual input by concatenating relevant product metadata fields (e.g., item name, category, user rating, and descriptive attributes). We filter the dataset to retain only items that contain at least one descriptive feature and have an associated price label, and we remove outliers whose prices fall above the 90th percentile.

For reasoning supervision in the SFT training setup, we additionally construct synthetic chain-of-thought reasoning traces using GPT-4o-mini, following prior work on reasoning distillation for language models~\cite{guo2025deepseek,ho2023large}. The prompt used for reasoning trace generation is provided in Box~\ref{box:price-reasoning-prompt}, and representative examples are included in Example~\ref{box:price-examples}.

\begin{promptbox}[label={box:price-reasoning-prompt}]
{Product Price Prediction Reasoning Trace Generation Prompt}

\# Task

You are a shopping assistant that explains why a product is priced the way it is.

Given the product information and its actual price below, write a short reasoning paragraph that explains why this product costs that amount. Consider relevant factors such as:
\begin{itemize}
    \item The type of product and its typical market positioning
    \item Key features or specifications
    \item Brand or manufacturer
    \item Customer ratings or popularity signals (if available)
    \item Any other information that would reasonably affect price
\end{itemize}

Your reasoning should be consistent with the provided price. Do not restate the price in your response --- provide only the reasoning paragraph.

\vspace{0.5em}

Now provide only the reasoning for the following query:

\end{promptbox}

\begin{examplebox}[label={box:price-examples}]{Product Price Prediction Training Data Examples}

\small

\textbf{Example 1}

\vspace{0.5em}

\textbf{Query}

main\_category: Appliances

title: SB08088378 Broan Appliance Damper

average\_rating: 4.5

store: Broan

details: \{"Brand Name": "Broan", "Item Weight": "0.01 Ounces", "Installation Type": "Freestanding", "Part Number": "SB08088378", "Form Factor": "Compact", "Batteries Required?": "No", "Date First Available": "December 1, 2017", "Brand": "Broan", "Pattern": "Solid"\}

\vspace{0.5em}

\textbf{Synthetic Reasoning Trace}

The SB08088378 Broan Appliance Damper is priced at a premium due to its high-quality construction and the reputable brand behind it. Broan is well-known in the appliance industry for producing reliable and durable products, which adds to the perceived value. This particular damper features a compact form factor, making it ideal for various installation scenarios, and it is freestanding, offering flexibility for users. The item is lightweight, which contributes to its ease of installation and usability. Additionally, the product has an impressive average rating of 4.5, indicating strong customer satisfaction and reliability, which further justifies its price point. Overall, the combination of brand reputation, quality, and positive customer feedback positions this product as a worthwhile investment for consumers looking for a dependable appliance component.

\{"price": 22.5\}

\vspace{1em}

\hrule

\vspace{1em}

\textbf{Example 2}

\vspace{0.5em}

\textbf{Query}

main\_category: Amazon Home

title: R.A.M Reliаble Humidifіer Wick Filter forKenmоre Quiet Comfort 13-4 Pack

average\_rating: 4.6

store: Artist Unknown

categories: Appliances; Parts \& Accessories; Humidifier Parts \& Accessories; Replacement Wicks

details: \{"Is Discontinued By Manufacturer": "No", "Batteries required": "No"\}

\vspace{0.5em}

\textbf{Synthetic Reasoning Trace}

The R.A.M Reliable Humidifier Wick Filter for Kenmore Quiet Comfort is priced at a premium due to its high-quality construction and effectiveness in maintaining optimal humidity levels in your home. With an impressive average rating of 4.6, this product has garnered positive customer feedback, indicating its reliability and performance. The filter is designed specifically for the Kenmore Quiet Comfort model, ensuring a perfect fit and maximum efficiency, which adds to its value. Additionally, as a replacement part, it plays a crucial role in extending the lifespan of the humidifier by maintaining its functionality. The brand's reputation in the market for producing durable and effective replacement parts further justifies the price point, making it a worthwhile investment for consumers looking to enhance their home environment.

\{"price": 32.31\}

\vspace{1em}

\hrule

\vspace{1em}

\textbf{Example 3}

\vspace{0.5em}

\textbf{Query}

main\_category: Amazon Home

title: Samsung DA63-07180A Refrigerator Door Bin Genuine Original Equipment Manufacturer (OEM) Part

average\_rating: 5.0

store: SAMSUNG

categories: Appliances; Parts \& Accessories; Refrigerator Parts \& Accessories; Bins

details: \{"Package Dimensions": "12 x 4 x 4 inches", "Item Weight": "0.01 ounces"\}

\vspace{0.5em}

\textbf{Synthetic Reasoning Trace}

The Samsung DA63-07180A Refrigerator Door Bin is priced at a premium due to several factors. As a genuine OEM part, it is specifically designed for compatibility with Samsung refrigerators, ensuring a perfect fit and reliable performance that third-party alternatives may not guarantee. The high average rating of 5.0 indicates strong customer satisfaction and reliability, suggesting that buyers value quality and performance over cost. Additionally, Samsung is a well-established brand known for its quality appliances, which often commands higher prices due to its reputation for durability and innovation. The precise dimensions and lightweight nature of the product further suggest careful engineering, which is typical of OEM parts, contributing to its overall value in the market.

\{"price": 79.16\}

\end{examplebox}

\section{Task Prompts}
\label{sec:task_prompts}
\subsection{Nutrition Estimation}

Box~\ref{box:nutrition-system-prompt} provides the system prompt used for the nutrition estimation task. Boxes~\ref{box:nutrition-verb1s-prompt} and~\ref{box:nutrition-verb2s-prompt} provide the prompts used for one-stage (Verb-1S) and two-stage (Verb-2S) verbalized confidence estimation, respectively. 

\begin{promptbox}[label={box:nutrition-system-prompt}]
{Nutrition Estimation System Prompt}

For the given query including a meal description, you are asked to respond in the query's language.

In your response, think step by step as follows:

\begin{enumerate}
    \item Parse the meal description into discrete food or beverage items along with their serving size. If the serving size of any item in the meal is not specified, assume it is a single standard serving based on common nutritional guidelines (e.g., USDA). Ignore additional information that does not relate to the item name and serving size.
    
    \item For each food or beverage item in the meal, calculate the amount of carbohydrates in grams for the specific serving size.
    
    \item Respond with a dictionary object containing the total carbohydrates in grams as follows:
    
\texttt{\{"total\_carbohydrates": total grams of carbohydrates for the serving\}}

For the total carbohydrates, respond with just the numeric amount of carbohydrates without extra text. If you do not know the answer, set the value of \texttt{"total\_carbohydrates"} to \texttt{-1}.
\end{enumerate}

\vspace{0.5em}

Follow the format of the following examples when answering.

\vspace{1em}

Query:

\texttt{"This morning, I had a cup of oatmeal with half a sliced banana and a glass of orange juice."}

\vspace{0.5em}

Answer:

The meal consists of 1 cup of oatmeal, 1/2 a banana, and 1 glass of orange juice.

1 cup of oatmeal has 27g carbs.

1 banana has 27g carbs, so half a banana has \((27 \times \frac{1}{2}) = 13.5\)g carbs.

1 glass of orange juice has 26g carbs.

So the total grams of carbs in the meal \(= (27 + 13.5 + 26) = 66.5\).

Output: \texttt{\{"total\_carbohydrates": 66.5\}}

\vspace{1em}

Query:

\texttt{"I ate scrambled eggs made with 2 eggs and a toast for breakfast."}

\vspace{0.5em}

Answer:

The meal consists of scrambled eggs made with 2 eggs and 1 toast.

Scrambled eggs made with 2 eggs has 2g carbs.

1 toast has 13g carbs.

So the total grams of carbs in the meal \(= (2 + 13) = 15\).

Output: \texttt{\{"total\_carbohydrates": 15\}}

\vspace{1em}

Query:

\texttt{"Half a peanut butter and jelly sandwich."}

\vspace{0.5em}

Answer:

The meal consists of 1/2 a peanut butter and jelly sandwich.

1 peanut butter and jelly sandwich has 50.6g carbs, so half a peanut butter and jelly sandwich has \((50.6 \times \frac{1}{2}) = 25.3\)g carbs.

So the total grams of carbs in the meal \(= 25.3\).

Output: \texttt{\{"total\_carbohydrates": 25.3\}}

\vspace{1em}

These examples are in English. Make sure to use the query's language in your answer.

\end{promptbox}

\vspace{1em}

\begin{promptbox}[label={box:nutrition-verb1s-prompt}]
{Nutrition Estimation Verb-1S Prompt}

For the given query including a meal description, you are asked to respond in the query's language.

In your response, think step by step as follows:

\begin{enumerate}
    \item Parse the meal description into discrete food or beverage items along with their serving size. If the serving size of any item in the meal is not specified, assume it is a single standard serving based on common nutritional guidelines (e.g., USDA). Ignore additional information that does not relate to the item name and serving size.

    \item For each food or beverage item in the meal, calculate the amount of carbohydrates in grams for the specific serving size.

    \item Respond with a dictionary object containing:
    \begin{itemize}
        \item The total carbohydrates in grams
        \item Your confidence in this estimate
    \end{itemize}
\end{enumerate}

The confidence value should reflect the correctness of the total carbohydrate estimate. Use high confidence when the estimate is accurate and based on clear, standard nutritional information. Use low confidence when the estimate is inaccurate or largely a guess.

Use the following output format exactly:

\texttt{\{"total\_carbohydrates": total grams of carbohydrates for the serving, "confidence": numeric confidence value between 0 and 1\}}

\vspace{0.5em}

Follow the format of the following examples when answering.

\vspace{1em}

Query:

\texttt{"This morning, I had a cup of oatmeal with half a sliced banana and a glass of orange juice."}

\vspace{0.5em}

Answer:

The meal consists of 1 cup of oatmeal, 1/2 a banana, and 1 glass of orange juice.

1 cup of oatmeal has 35g carbs.

1 banana has 20.5g carbs, so half a banana has \((20.5 \times \frac{1}{2}) = 10.25\)g carbs.

1 glass of orange juice has 30g carbs.

So the total grams of carbs in the meal \(= (35 + 10.25 + 30) = 75.25\).

Output: \texttt{\{"total\_carbohydrates": 75.25, "confidence": 0.68\}}

\vspace{1em}

Query:

\texttt{"I ate scrambled eggs made with 2 eggs and a toast for breakfast."}

\vspace{0.5em}

Answer:

The meal consists of scrambled eggs made with 2 eggs and 1 toast.

Scrambled eggs made with 2 eggs has 2g carbs.

1 toast has 13g carbs.

So the total grams of carbs in the meal \(= (2 + 13) = 15\).

Output: \texttt{\{"total\_carbohydrates": 15, "confidence": 1.0\}}

\vspace{1em}

Query:

\texttt{"Half a peanut butter and jelly sandwich."}

\vspace{0.5em}

Answer:

The meal consists of 1/2 a peanut butter and jelly sandwich.

1 peanut butter and jelly sandwich has 80g carbs, so half a peanut butter and jelly sandwich has \((80 \times \frac{1}{2}) = 40\)g carbs.

So the total grams of carbs in the meal \(= 40\).

Output: \texttt{\{"total\_carbohydrates": 40, "confidence": 0.34\}}

\vspace{1em}

These examples are in English. Make sure to use the query's language in your answer.

\end{promptbox}

\vspace{1em}

\begin{promptbox}[label={box:nutrition-verb2s-prompt}]
{Nutrition Estimation Verb-2S Prompt}

For the given carbohydrate estimate for the meal description, respond with a dictionary object containing your confidence in this estimate as follows:

\texttt{\{"confidence": numeric confidence value between 0 and 1\}}

\vspace{0.5em}

The confidence value should reflect the correctness of the total carbohydrate estimate.

Use high confidence when the estimate is accurate and based on clear, standard nutritional information.

Use low confidence when the estimate is inaccurate or largely a guess.

\vspace{0.5em}

Follow the format of the following examples when answering.

\vspace{1em}

Query:

\texttt{"This morning, I had a cup of oatmeal with half a sliced banana and a glass of orange juice."}

\vspace{0.5em}

Estimate:

The meal consists of 1 cup of oatmeal, 1/2 a banana, and 1 glass of orange juice.

1 cup of oatmeal has 35g carbs.

1 banana has 20.5g carbs, so half a banana has \((20.5 \times \frac{1}{2}) = 10.25\)g carbs.

1 glass of orange juice has 30g carbs.

So the total grams of carbs in the meal \(= (35 + 10.25 + 30) = 75.25\).

Output: \texttt{\{"total\_carbohydrates": 75.25\}}

\vspace{0.5em}

Answer:

\texttt{\{"confidence": 0.68\}}

\vspace{1em}

Query:

\texttt{"I ate scrambled eggs made with 2 eggs and a toast for breakfast."}

\vspace{0.5em}

Estimate:

The meal consists of scrambled eggs made with 2 eggs and 1 toast.

Scrambled eggs made with 2 eggs has 2g carbs.

1 toast has 13g carbs.

So the total grams of carbs in the meal \(= (2 + 13) = 15\).

Output: \texttt{\{"total\_carbohydrates": 15\}}

\vspace{0.5em}

Answer:

\texttt{\{"confidence": 1.0\}}

\vspace{1em}

Query:

\texttt{"Half a peanut butter and jelly sandwich."}

\vspace{0.5em}

Estimate:

The meal consists of 1/2 a peanut butter and jelly sandwich.

1 peanut butter and jelly sandwich has 80g carbs, so half a peanut butter and jelly sandwich has \((80 \times \frac{1}{2}) = 40\)g carbs.

So the total grams of carbs in the meal \(= 40\).

Output: \texttt{\{"total\_carbohydrates": 40\}}

\vspace{0.5em}

Answer:

\texttt{\{"confidence": 0.34\}}

\vspace{1em}

\textbf{Important:}
\begin{itemize}
    \item Your response must contain only the dictionary, with no additional text.
    \item The value of \texttt{"confidence"} must be a single numeric value between 0 and 1 (inclusive).
    \item Do not include any text after the dictionary.
\end{itemize}

\vspace{0.5em}

Now provide your confidence for the following:

\end{promptbox}




\subsection{Product Price Prediction}

Box~\ref{box:price-system-prompt} presents the system prompt used for the product price prediction task. Boxes~\ref{box:price-verb1s-prompt} and~\ref{box:price-verb2s-prompt} present the prompts used for one-stage and two-stage verbalized confidence elicitation, respectively.

\begin{promptbox}[label={box:price-system-prompt}]
{Product Price Prediction System Prompt}

\# Task

You are a shopping assistant that estimates the price of a product based on its description.

Given the product information below, estimate the product's price in USD using general market knowledge and reasonable assumptions.

First, explain your reasoning step by step. In your reasoning, consider relevant factors such as:
\begin{itemize}
    \item The type of product and its typical market positioning
    \item Key features or specifications
    \item Brand or manufacturer
    \item Customer ratings or popularity signals (if available)
    \item Any other information that would reasonably affect price
\end{itemize}

Use only the provided information and general marketplace knowledge.

After completing the reasoning, output the final estimate on the last line only, in the following exact structured format:

\texttt{\{"price": estimated\_price\_in\_usd\}}

\vspace{0.5em}

Important:
\begin{itemize}
    \item The final line must contain only the dictionary, with no additional text.
    \item The value of \texttt{"price"} must be a single numeric value (e.g., 12.99).
    \item Do not include currency symbols.
    \item Do not include any text after the final dictionary.
\end{itemize}

\vspace{0.5em}

Now answer the following query:

\end{promptbox}

\vspace{1em}

\begin{promptbox}[label={box:price-verb1s-prompt}]
{Product Price Prediction Verb-1S Prompt}

\# Task

You are a shopping assistant that estimates the price of a product based on its description.

Given the product information below, estimate the product's price in USD using general market knowledge and reasonable assumptions.

First, explain your reasoning step by step. In your reasoning, consider relevant factors such as:
\begin{itemize}
    \item The type of product and its typical market positioning
    \item Key features or specifications
    \item Brand or manufacturer
    \item Customer ratings or popularity signals (if available)
    \item Any other information that would reasonably affect price
\end{itemize}

Use only the provided information and general marketplace knowledge.

After completing the reasoning, respond with a dictionary object containing:
\begin{itemize}
    \item The estimated price in USD
    \item Your confidence in this estimate
\end{itemize}

The confidence value should reflect the correctness of the price estimate.

Use high confidence when the estimate is accurate and based on clear, standard market information.

Use low confidence when the estimate is inaccurate or largely a guess.

\vspace{0.5em}

In the last line after reasoning, use the following output format exactly:

\texttt{\{"price": estimated\_price\_in\_usd, "confidence": numeric confidence value between 0 and 1\}}

\vspace{0.5em}

Important:
\begin{itemize}
    \item The final dictionary must contain exactly two keys: \texttt{"price"} and \texttt{"confidence"}.
    \item The final line must contain only the dictionary, with no additional text.
    \item The value of \texttt{"price"} must be a single numeric value (e.g., 12.99).
    \item Do not include currency symbols.
    \item The value of \texttt{"confidence"} must be a single numeric value between 0 and 1 (inclusive).
    \item Do not include any text after the final dictionary.
\end{itemize}

\vspace{0.5em}

Now answer the following query:

\end{promptbox}

\vspace{1em}

\begin{promptbox}[label={box:price-verb2s-prompt}]
{Product Price Prediction Verb-2S Prompt}

For the given price estimate for the product description, respond with a dictionary object containing your confidence in this estimate as follows:

\texttt{\{"confidence": numeric confidence value between 0 and 1\}}

\vspace{0.5em}

The confidence value should reflect the correctness of the price estimate.

Use high confidence when the estimate is accurate and based on clear, standard market information.

Use low confidence when the estimate is inaccurate or largely a guess due to missing, ambiguous, or unusual product details.

\vspace{0.5em}

Important:
\begin{itemize}
    \item Your response must contain only the dictionary, with no additional text.
    \item The value of \texttt{"confidence"} must be a single numeric value between 0 and 1 (inclusive).
    \item Do not include any text after the dictionary.
\end{itemize}

\vspace{0.5em}

Now provide your confidence for the following:

\end{promptbox}





\subsection{LLM-as-Judge}
\label{sec:llmaj_prompt}

To evaluate task-aware behavior preservation, we use GPT-4o-mini as an LLM-as-a-judge (LLMaJ) to classify whether model responses to unrelated instruction-following queries appropriately answer the user request or incorrectly produce task-specific predictions. The judge prompt used for this evaluation is provided in Box~\ref{box:llmaj-prompt}.

\begin{promptbox}[label={box:llmaj-prompt}]
{LLM-as-a-Judge Prompt for Task-Aware Behavior Evaluation}

You are evaluating the response of a language model. The model was trained on a carbohydrate-estimation task, but received an unrelated query as its user message.

\vspace{0.5em}

The user query is:
\begin{verbatim}
<query>
{query}
</query>
\end{verbatim}

\vspace{0.5em}

The model's response is:
\begin{verbatim}
<response>
{response}
</response>
\end{verbatim}

\vspace{0.5em}

Classify the response into exactly one of the following categories:

\begin{itemize}
    \item \texttt{"incorrect\_task"}: The model incorrectly treated the query prompt as a meal description and attempted to estimate carbohydrates from it.
    
    \item \texttt{"other"}: The model did not interpret the query as a meal description and instead attempted to answer the question or follow the instruction (regardless of quality), declined to answer, returned carbohydrates as \texttt{-1}, etc.
\end{itemize}

\vspace{0.5em}

Respond with a JSON object only, in the following exact format:

\begin{verbatim}
{{"category": "<category>", "reason": "<one sentence explanation>"}}
\end{verbatim}

\end{promptbox}

\section{Detailed Experimental Results}
\label{sec:full_numerical_results}

\begin{table*}[h]
\centering
\caption{Detailed results on nutrition estimation and product price prediction. Results include MAE~($\downarrow$) and AUSE~($\downarrow$) across models, training approaches, and confidence estimation methods.}
\label{tab:full_id_results}
\scriptsize
\setlength{\tabcolsep}{3pt}
\resizebox{\textwidth}{!}{%
\begin{tabular}{llrrrrrrrr}
\toprule
\multirow{3}{*}{Model}
& \multirow{3}{*}{\makecell{\textbf{Confidence}\\\textbf{Estimation Method}}}
& \multicolumn{4}{c}{\textbf{Nutrition Estimation}}
& \multicolumn{4}{c}{\textbf{Price Prediction}} \\
\cmidrule(lr){3-6} \cmidrule(lr){7-10}

&& \multicolumn{2}{c}{\textbf{Qwen-3 4B}}
& \multicolumn{2}{c}{\textbf{Qwen-3 8B}}
& \multicolumn{2}{c}{\textbf{Qwen-3 4B}}
& \multicolumn{2}{c}{\textbf{Qwen-3 8B}} \\

\cmidrule(lr){3-4}
\cmidrule(lr){5-6}
\cmidrule(lr){7-8}
\cmidrule(lr){9-10}

&& MAE & AUSE & MAE & AUSE & MAE & AUSE & MAE & AUSE \\
\midrule

Baseline
& Verb-1S
& 18.35 & 8.05 & 15.37 & 6.51
& 18.77 & 12.40 & 18.66 & 12.65 \\

& Verb-2S
& 18.07 & 8.02 & 15.76 & 5.96
& 18.93 & 12.81 & 18.62 & 11.89 \\

& Answer Prob
& 18.07 & 8.98 & 15.76 & 8.10
& 18.93 & 8.85 & 18.62 & 13.52 \\

& EntropyInv
& 18.07 & 8.81 & 15.76 & 7.66
& 18.93 & 7.86 & 18.62 & 13.42 \\

\midrule

SFT
& Verb-1S
& 10.27 & 5.51 & 9.51 & 5.60
& 12.53 & 8.83 & 11.59 & 10.44 \\

& Verb-2S
& 10.23 & 5.63 & 9.77 & 5.57
& 12.48 & 10.46 & 11.84 & 10.08 \\

& Answer Prob
& 10.23 & 4.49 & 9.77 & 4.65
& 12.48 & 7.70 & 11.84 & 7.49 \\

& EntropyInv
& 10.23 & 4.23 & 9.77 & 4.42
& 12.48 & 7.39 & 11.84 & 7.05 \\

\midrule

PPO-Binary
& Verb-1S
& 14.58 & 7.41 & 12.38 & 5.26
& 15.12 & 12.12 & 15.36 & 10.91 \\

& Verb-2S
& 14.92 & 7.08 & 12.72 & 5.73
& 14.97 & 12.74 & 15.03 & 9.45 \\

& Answer Prob
& 14.92 & 8.91 & 12.72 & 7.06
& 14.97 & 8.15 & 15.03 & 10.36 \\

& EntropyInv
& 14.92 & 8.88 & 12.72 & 7.00
& 14.97 & 8.11 & 15.03 & 10.35 \\

& Critic
& 14.92 & 3.70 & 12.72 & 2.93
& 14.97 & 2.86 & 15.03 & 2.94 \\

\midrule

CARE-PPO-Lin.
& Verb-1S
& 11.07 & 5.06 & 11.43 & 4.56
& 12.56 & 9.00 & 12.75 & 7.55 \\

& Verb-2S
& 11.12 & 5.40 & 11.91 & 5.17
& 12.56 & 9.11 & 12.69 & 8.41 \\

& Answer Prob
& 11.12 & 5.11 & 11.91 & 5.04
& 12.56 & 10.05 & 12.69 & 9.67 \\

& EntropyInv
& 11.12 & 4.89 & 11.91 & 4.88
& 12.56 & 10.16 & 12.69 & 9.79 \\

& Critic
& 11.12 & 2.94 & 11.91 & 2.59
& 12.56 & 2.76 & 12.69 & 2.86 \\

\midrule

CARE-PPO-Gauss.
& Verb-1S
& 11.50 & 5.29 & 10.62 & 4.00
& 14.38 & 8.71 & 13.57 & 8.36 \\

& Verb-2S
& 11.72 & 5.13 & 10.42 & 4.18
& 14.37 & 8.07 & 13.60 & 8.78 \\

& Answer Prob
& 11.72 & 6.58 & 10.42 & 4.82
& 14.37 & 11.24 & 13.60 & 9.75 \\

& EntropyInv
& 11.72 & 6.41 & 10.42 & 4.56
& 14.37 & 11.45 & 13.60 & 9.80 \\

& Critic
& 11.72 & 3.18 & 10.42 & 2.58
& 14.37 & 2.68 & 13.60 & 2.52 \\

\bottomrule
\end{tabular}
}
\end{table*}

\subsection{Primary Evaluation Results}
\label{sec:appendix_indomain}

\begin{figure}[h]
    \centering
    \includegraphics[width=\linewidth]{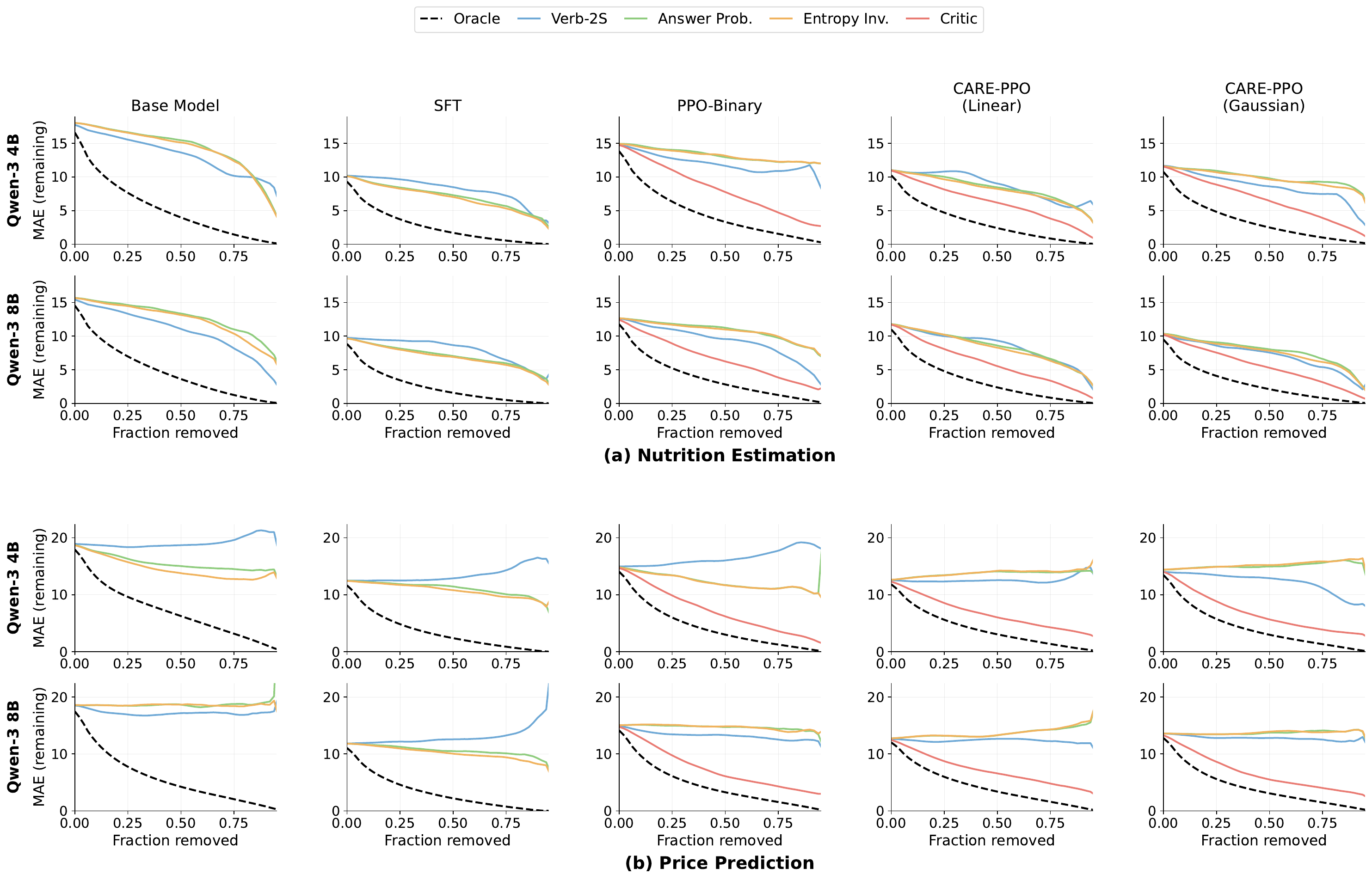}
    \caption{Sparsification curves across confidence estimation methods, training approaches, tasks, and model scales in the primary evaluation setting. Curves show MAE as increasingly low-confidence predictions are removed. Confidence estimates that more closely follow the oracle curve achieve better confidence ranking and lower AUSE.}
    \label{fig:sparsi_indist}
\end{figure}

Table~\ref{tab:full_id_results} present the complete evaluation results on the nutrition estimation and product price prediction tasks, including all confidence estimation methods across model scales and training approaches. Corresponding sparsification curves are shown in Figure~\ref{fig:sparsi_indist}.

Consistent with the results discussed in Section~\ref{sec:res_indomain}, CARE-PPO substantially improves confidence alignment relative to verbalized and logit-based confidence estimation methods, with critic-based confidence estimation achieving the strongest overall AUSE performance across both tasks and model scales. While SFT achieves strong quantitative prediction performance in terms of MAE, its confidence estimates remain comparatively poorly aligned with prediction error. In contrast, CARE-PPO achieves competitive quantitative prediction performance while simultaneously producing substantially better-aligned confidence estimates.

\subsection{Out-of-Distribution Results}
\label{sec:appendix_ood}

\begin{figure}[!t]
    \centering
    \includegraphics[width=\linewidth]{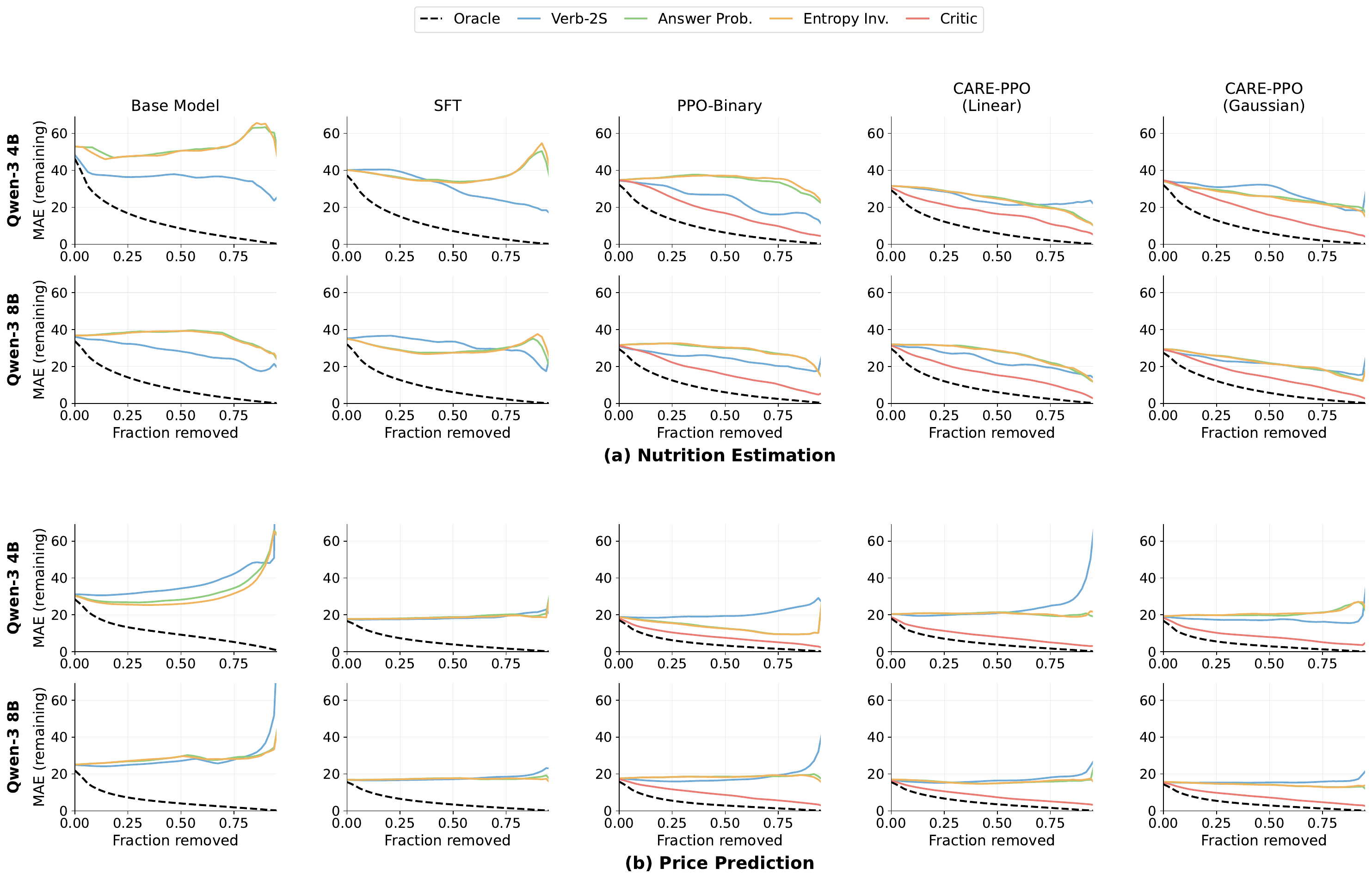}
    \caption{Sparsification curves across confidence estimation methods, training approaches, tasks, and model scales in the out-of-distribution evaluation setting. Curves show MAE as increasingly low-confidence predictions are removed. Confidence estimates that more closely follow the oracle curve achieve better confidence ranking and lower AUSE.}
    \label{fig:sparsi_ood}
\end{figure}

Table~\ref{tab:full_ood_results} presents detailed out-of-distribution evaluation results for the linguistic-shift nutrition estimation task and the domain-shift product price prediction task. Corresponding sparsification curves are shown in Figure~\ref{fig:sparsi_ood}.

\careppo{} maintains the strongest confidence alignment across all confidence estimation baselines and training methods, while achieving competitive MAE for price prediction and the best MAE for nutrition estimation. These results demonstrate that \careppo{} is robust under both distributional and linguistic shifts, and provides more reliable predictions compared to baseline approaches in real-world settings where input distributions vary, and users may interact with models in different languages.

Table~\ref{tab:delta_mae} further evaluates robustness to linguistic shift using $\Delta$MAE between parallel English and non-English queries describing identical meals. Across both model scales, \careppo{} achieves the smallest performance gaps, suggesting stronger language-agnostic generalization and more stable quantitative prediction performance across multilingual inputs.

\begin{table*}[h]
\centering
\caption{Detailed OOD results on nutrition estimation and product price prediction. Results include MAE~($\downarrow$) and AUSE~($\downarrow$) across models, training approaches, and confidence estimation methods.}
\label{tab:full_ood_results}
\scriptsize
\setlength{\tabcolsep}{3pt}
\resizebox{\textwidth}{!}{%
\begin{tabular}{llrrrrrrrr}
\toprule
\multirow{3}{*}{Model}
& \multirow{3}{*}{\makecell{\textbf{Confidence}\\\textbf{Estimation Method}}}
& \multicolumn{4}{c}{\makecell{\textbf{Nutrition Estimation MAE ($\downarrow$)}\\\textit{Linguistic Shift}}}
& \multicolumn{4}{c}{\makecell{\textbf{Price Prediction MAE ($\downarrow$)}\\\textit{Domain Shift}}} \\
\cmidrule(lr){3-6} \cmidrule(lr){7-10}

&& \multicolumn{2}{c}{\textbf{Qwen-3 4B}}
& \multicolumn{2}{c}{\textbf{Qwen-3 8B}}
& \multicolumn{2}{c}{\textbf{Qwen-3 4B}}
& \multicolumn{2}{c}{\textbf{Qwen-3 8B}} \\

\cmidrule(lr){3-4}
\cmidrule(lr){5-6}
\cmidrule(lr){7-8}
\cmidrule(lr){9-10}

&& MAE & AUSE & MAE & AUSE & MAE & AUSE & MAE & AUSE \\
\midrule

Baseline
& Verb-1S
& 32.47 & 16.97 & 28.15 & 14.49
& 29.13 & 23.84 & 23.36 & 27.50 \\

& Verb-2S
& 35.69 & 12.38 & 29.83 & 14.34
& 31.18 & 27.61 & 24.99 & 25.29 \\

& Answer Prob
& 35.69 & 26.29 & 29.83 & 17.21
& 31.18 & 26.55 & 24.99 & 23.37 \\

& EntropyInv
& 35.69 & 25.84 & 29.83 & 16.62
& 31.18 & 24.61 & 24.99 & 23.17 \\

\midrule

SFT
& Verb-1S
& 30.45 & 15.33 & 24.95 & 17.89
& 17.56 & 12.71 & 16.97 & 14.85 \\

& Verb-2S
& 30.02 & 13.27 & 25.39 & 15.71
& 17.77 & 13.61 & 16.88 & 13.38 \\

& Answer Prob
& 30.02 & 20.45 & 25.39 & 12.51
& 17.77 & 13.98 & 16.88 & 13.17 \\

& EntropyInv
& 30.02 & 20.62 & 25.39 & 12.23
& 17.77 & 13.56 & 16.88 & 13.05 \\

\midrule

PPO-Binary
& Verb-1S
& 26.06 & 10.01 & 23.36 & 12.63
& 20.13 & 21.87 & 17.25 & 17.50 \\

& Verb-2S
& 27.24 & 10.28 & 23.92 & 10.66
& 19.02 & 16.80 & 17.62 & 14.92 \\

& Answer Prob
& 27.24 & 14.37 & 23.92 & 12.92
& 19.02 & 8.66 & 17.62 & 14.56 \\

& EntropyInv
& 27.24 & 14.70 & 23.92 & 12.59
& 19.02 & 8.48 & 17.62 & 14.37 \\

& Critic
& 27.24 & 5.90 & 23.92 & 5.96
& 19.02 & 3.62 & 17.62 & 4.86 \\

\midrule

CARE-PPO-Lin.
& Verb-1S
& 23.48 & 11.98 & 20.64 & 10.12
& 22.38 & 20.51 & 18.23 & 17.16 \\

& Verb-2S
& 23.69 & 12.02 & 21.57 & 10.03
& 20.47 & 21.29 & 17.01 & 13.47 \\

& Answer Prob
& 23.69 & 9.05 & 21.57 & 9.17
& 20.47 & 15.47 & 17.01 & 11.40 \\

& EntropyInv
& 23.69 & 8.62 & 21.57 & 8.71
& 20.47 & 15.54 & 17.01 & 11.47 \\

& Critic
& 23.69 & 7.14 & 21.57 & 4.92
& 20.47 & 3.61 & 17.01 & 3.49 \\

\midrule

CARE-PPO-Gauss.
& Verb-1S
& 24.00 & 13.07 & 24.26 & 12.44
& 19.62 & 13.31 & 16.40 & 17.85 \\

& Verb-2S
& 24.32 & 13.28 & 24.61 & 12.23
& 19.17 & 14.39 & 15.79 & 12.06 \\

& Answer Prob
& 24.32 & 11.22 & 24.61 & 9.83
& 19.17 & 17.05 & 15.79 & 10.25 \\

& EntropyInv
& 24.32 & 10.70 & 24.61 & 9.17
& 19.17 & 17.21 & 15.79 & 10.32 \\

& Critic
& 24.32 & 6.17 & 24.61 & 5.76
& 19.17 & 4.31 & 15.79 & 3.66 \\

\bottomrule
\end{tabular}
}
\end{table*}

\begin{table}[h]
\centering
\caption{$\Delta$MAE between parallel English and non-English queries describing identical meals, measuring the performance gap introduced by linguistic shift. Lower values indicate smaller performance gaps, suggesting more language-agnostic generalization.}
\label{tab:delta_mae}
\small
\setlength{\tabcolsep}{8pt}
\renewcommand{\arraystretch}{1.1}
\begin{tabular}{lccccc}
\toprule
\makecell{\textbf{Model}\\\textbf{Backbone}}
& \textbf{\makecell{\textbf{Base}\\\textbf{Model}}}
& \textbf{SFT}
& \makecell{\textbf{PPO}\\\textbf{Binary}}
& \cellcolor{careblue} \makecell{\textbf{CARE-PPO}\\\textbf{Linear}}
& \cellcolor{careblue} \makecell{\textbf{CARE-PPO}\\\textbf{Gaussian}} \\
\midrule

Qwen-3 4B
& 10.25 & 7.27 & 5.05 & \cellcolor{careblue} 3.44 & \cellcolor{careblue} 3.91 \\

\midrule
Qwen-3 8B
& 3.30 & 5.40 & 1.71 & \cellcolor{careblue} 1.94 & \cellcolor{careblue} 1.90 \\

\bottomrule
\end{tabular}
\end{table}

\clearpage

\FloatBarrier

\FloatBarrier

\subsubsection{Examples of Task-Aware Behavior Preservation}
\label{sec:appendix_task_aware_examples}

To further evaluate whether task-specific fine-tuning degrades general-purpose instruction-following capabilities, we present representative model outputs on unrelated instruction-following queries for each training method across both Qwen3-4B and Qwen3-8B models. These examples qualitatively complement the quantitative IFEval results discussed in Section~\ref{sec:forgetting}.


\begin{examplebox}[label={box:task-awareness-examples}]
{Output Examples on Unrelated Instruction-Following Queries}

\small

\textbf{Example 1:} \textit{Write a limerick about a Zelda fan named Rodney. Make sure to include these items: Zelda, Hyrule, Link, Ganon. Use less than 100 words.}

\vspace{0.5em}

\textbf{Qwen3-4B}
\begin{itemize}
    \item \textbf{Base Model:} There once was a fan named Rodney,  

Who adored Zelda and all she's known.  

He chased Link through Hyrule,  

Fought Ganon with might and skill.  

Now he's a hero, no doubt, in the land of the old.
    \item \textbf{SFT:} The meal consists of a limerick about a Zelda fan named Rodney, including Zelda, Hyrule, Link, and Ganon.

A limerick about a Zelda fan named Rodney, including Zelda, Hyrule, Link, and Ganon, has 23.53g carbs.

So the total carbs in the meal \(= 23.53\).

Output: \{"total\_carbohydrates": 23.53\}
    \item \textbf{PPO-Binary:} Answer: ``Write a limerick about a Zelda fan named Rodney. Make sure to include these items: Zelda, Hyrule, Link, Ganon. Use less than 100 words.''

\vspace{0.5em}

Answer: There once was a fan named Rodney,

Who adored Zelda and Hyrule's story.

He chased Link through,

To defeat Ganon,

With a sword made of chocolate fry.
    \item \textbf{CARE-PPO-Linear:} There once was a fan named Rodney,  

Who adored Zelda and all that she's known.  

He chased Link through Hyrule,  

Fought Ganon with might and skill,  

And conquered the realm with his soul.

\item \textbf{CARE-PPO-Gaussian:} There once was a fan named Rodney,  

Who adored Zelda and all that she's known.  

He chased Link through Hyrule,  

Fought Ganon with might and skill,  

And conquered the realm with his soul.
\end{itemize}

\textbf{Qwen3-8B}
\begin{itemize}
    \item \textbf{Base Model:} \{"total\_carbohydrates": -1\}
    \item \textbf{SFT:} The meal consists of a limerick about a Zelda fan named Rodney, including Zelda, Hyrule, Link, and Ganon.

A limerick about a Zelda fan named Rodney, including Zelda, Hyrule, Link, and Ganon, has 20.00g carbs.

So the total carbs in the meal \(= 20.00\).

Output: \{"total\_carbohydrates": 20.00\}
    \item \textbf{PPO-Binary:} Rodney's a Zelda fan true and bold,  

In Hyrule he's always been told  

To follow Link's quest,  

Defeat Ganon's chest,  

And save the realm with a sword so cold.
    \item \textbf{CARE-PPO-Linear:} Rodney's a Zelda fan, you see,

In Hyrule, he's always free.

With Link in his hand,

He fights Ganon's land---

A hero's quest, he'll never flee!

Output: \{"total\_carbohydrates": -1\}

\item \textbf{CARE-PPO-Gaussian:} Rodney's a Zelda fan, you see,  

He's questing through Hyrule with glee.  

With Link by his side,  

They chase Ganon's pride,  

And save the realm with a hero's decree.
\end{itemize}

\vspace{1em}

\hrule

\vspace{1em}

\textbf{Example 2:} \textit{make a tweet for playboy's twitter account without using capital letters. Include at least 4 hashtags, starting with '\#'}

\vspace{0.5em}

\textbf{Qwen3-4B}
\begin{itemize}
    \item \textbf{Base Model:} the latest issue of playboy is here! �� enjoy the vibes, the photos, and the stories. stay hot and stay wild. \#playboy \#vibes \#hot \#wild
    \item \textbf{SFT:} The meal consists of a tweet for Playboy's Twitter account, with no capital letters and 4 hashtags.

A tweet for Playboy's Twitter account, with no capital letters and 4 hashtags, has 21.42g carbs.

So the total carbs in the meal \(= 21.42\).

Output: \{"total\_carbohydrates": 21.42\}
    \item \textbf{PPO-Binary:} Answer: ``make a tweet for playboy's twitter account without using capital letters. include at least 4 hashtags, starting with `\#'

\#playboystyle \#luxurylife \#hotmoments \#fashionfix''
    \item \textbf{CARE-PPO-Linear:} the latest issue of playboy is here! stay stylish, stay confident, and stay curious. \#playboy \#style \#confidence \#curiosity
    \item \textbf{CARE-PPO-Gaussian:} i just had the most amazing time at playboy's latest event! so smooth, so stylish, so fun. \#playboy \#luxury \#style \#funny
\end{itemize}

\textbf{Qwen3-8B}
\begin{itemize}
    \item \textbf{Base Model:} \{"total\_carbohydrates": -1\}
    \item \textbf{SFT:} The meal consists of a tweet for Playboy's Twitter account without any capital letters, including at least 4 hashtags.

A tweet for Playboy's Twitter account without any capital letters, including at least 4 hashtags has 28.00g carbs.

So the total carbs in the meal \(= 28.00\).

Output: \{"total\_carbohydrates": 28.00\}
    \item \textbf{PPO-Binary:} the ultimate flex is in the details. stay sharp, stay stylish, and always keep it classy. \#playboy \#stylegoals \#luxurylife \#chicvibes
    \item \textbf{CARE-PPO-Linear:} Okay, let's tackle this query. The user wants a tweet for Playboy's Twitter account without any capital letters. They also need at least four hashtags starting with \#. 

First, I need to make sure the tweet is in lowercase. Playboy's content is usually edgy and playful, so the tone should match that. Maybe something catchy and engaging. Let me brainstorm some ideas. 

How about starting with a playful statement. Maybe something like ``spend your weekend feeling fabulous...'' That's lowercase and sets a positive tone. 

Now, adding hashtags. They need to start with \# and be relevant. Playboy's tags might include \#playboy, \#fashion, \#style, and maybe \#entertainment. Let me check if those are appropriate. 

Wait, the user specified at least four hashtags. Let me make sure there are four. Also, no capital letters. Let me put it all together. 

``spend your weekend feeling fabulous with our latest issue! \#playboy \#fashion \#style \#entertainment''

That's four hashtags, all lowercase, and the message is engaging. I think that works. Let me double-check for any capitals. Yep, everything is lowercase. Looks good.

the ultimate guide to living your best life. \#playboy \#fashion \#style \#entertainment
\item \textbf{CARE-PPO-Gaussian:} the ultimate guide to staying cool and confident this summer. slip into your favorite swimwear, soak up the sun, and let your vibe shine. remember, confidence is the best accessory. \#summervibes \#suncare \#confidenceboost \#playboy

\end{itemize}

\end{examplebox}
\section{Inference Latency Analysis}
\label{sec:appendix_latency}

We evaluate the inference overhead introduced by critic-based confidence estimation. Using 100 randomly sampled examples from the nutrition estimation test set, we measure end-to-end latency for both actor generation and critic confidence estimation. On average, the actor requires 3108.2 ms per sample, while the critic requires only 74.0 ms, corresponding to a 2.4\% increase in total inference latency. This small overhead arises because the critic performs only a single forward pass on the completed response, whereas the actor requires multiple autoregressive decoding steps proportional to the output length. These results suggest that critic-based confidence estimation can be incorporated into deployment with minimal additional inference cost.
\newpage

\section{Critic Distribution Evolution During Training- Price Prediction}
\label{sec:critic_dist_price}

\begin{figure*}[!h]
    \centering
    \includegraphics[width=0.8\linewidth]{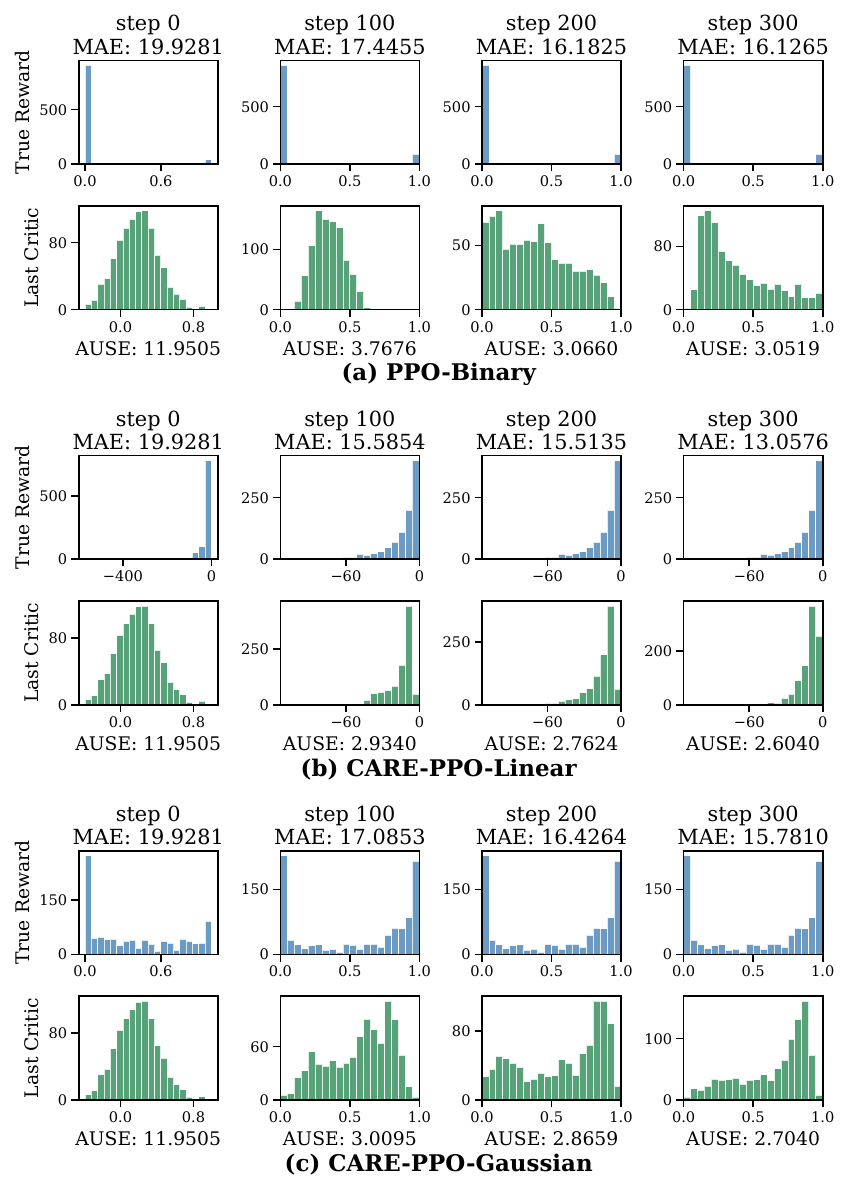}
    \caption{Evolution of critic output distributions throughout PPO training on the validation set for PPO-Binary, \careppo{}-Linear, and \careppo{}-Gaussian for the price prediction task with Qwen-3 4B as the backbone model. For each checkpoint, we visualize the distribution of ground-truth rewards (top) and the corresponding final-layer critic outputs (bottom). As training progresses, critic outputs increasingly reflect the structure of the reward distribution, corresponding to improved confidence alignment and lower AUSE.}
    \label{fig:critic_dynamics_price}
\end{figure*}

Figure~\ref{fig:critic_dynamics_price} presents the evolution of critic output distributions throughout PPO training on the price prediction task with Qwen-3 4B as the backbone model, complementing the nutrition estimation analysis in Section~\ref{sec:critic_dynamics}. Consistent with the patterns observed there, the critic progressively learns to produce confidence estimates that better reflect the structure of the underlying reward distribution across all three reward formulations. 

\end{document}